\documentclass[10pt,twocolumn,letterpaper]{article}

\usepackage{3dv}
\usepackage{times}
\usepackage{epsfig}
\usepackage{graphicx}
\usepackage{amsmath,amsfonts,amssymb}
\usepackage{color}
\usepackage{caption}
\usepackage{subcaption}
\usepackage{algorithm}
\usepackage{ntheorem}
\usepackage{algpseudocode}
\usepackage{booktabs}
\usepackage{multirow}
\usepackage{siunitx}%
\usepackage{xspace}
\usepackage{array}
\usepackage{cite}
\usepackage{booktabs}
\usepackage{multirow}
\usepackage{siunitx}
\usepackage{thmtools}
\usepackage[utf8]{inputenc}
\usepackage{adjustbox}
\usepackage[titletoc]{appendix}

\usepackage{microtype}
\graphicspath{{gfx/}}

\newcommand{\bI}{\mathbf{I}}

\newcommand{\bv}{\mathbf{v}}
\newcommand{\bq}{\mathbf{q}}
\newcommand{\bR}{\mathbf{R}}

\newcommand{\bp}{\mathbf{p}}
\newcommand{\bt}{\mathbf{t}}

\newcommand{\bP}{\mathbf{P}}

\newcommand{\bo}{\mathbf{o}}

\newcommand{\br}{\mathbf{r}}

\newcommand{\nR}{\mathbb{R}}

\newcommand{\cL}{\mathcal{L}}

\newcommand{\figref}[1]{\Fig~\ref{#1}}

\renewcommand{\eqref}[1]{Eq.~\ref{#1}}

\renewcommand{\vec}[1]{{\mathbf #1}}

\makeatletter
\newcommand{\raisemath}[1]{\mathpalette{\raisem@th{#1}}}
\newcommand{\raisem@th}[3]{\raisebox{#1}{$#2#3$}}
\makeatother

\makeatletter
\DeclareRobustCommand\onedot{\futurelet\@let@token\@onedot}
\def\@onedot{\ifx\@let@token.\else.\null\fi\xspace}
\def\eg{e.g\onedot} 
\def\ie{i.e\onedot}

\def\wrt{with respect to }
\def\etal{et~al\onedot} 
\def\Fig{Fig\onedot}   
\makeatother

\newcommand{\boldparagraph}[1]{\vspace{0.2cm}\noindent{\bf #1:} }
\newcommand{\boldtext}[1]{\vspace{0.2cm}\noindent{\bf #1} }

\definecolor{darkgreen}{rgb}{0,0.7,0}

\definecolor{darkyellow}{rgb}{0.8,0.8,0}
\definecolor{darkorange}{rgb}{0.8,0.5,0}
\definecolor{darkred}{rgb}{0.7,0,0}
\definecolor{darkpurple}{rgb}{0.8,0,0.8}
\definecolor{darkblue}{rgb}{0,0,1.0}
\definecolor{darkcyan}{rgb}{0,1.0,1.0}
\newcommand{\boldyellow}[1]{{\color{darkyellow}{\bf #1}}}
\newcommand{\boldgreen}[1]{{\color{darkgreen}{\bf #1}}}
\newcommand{\boldorange}[1]{{\color{darkorange}{\bf #1}}}
\newcommand{\boldred}[1]{{\color{darkred}{\bf #1}}}
\newcommand{\boldpurple}[1]{{\color{darkpurple}{\bf #1}}}
\newcommand{\boldblue}[1]{{\color{darkblue}{\bf #1}}}
\newcommand{\boldcyan}[1]{{\color{darkcyan}{\bf #1}}}

\theoremnumbering{arabic}
\newtheorem{theorem}{Theorem}

\newtheorem{claim}[theorem]{Claim}
\theoremstyle{nonumberplain}
\theoremsymbol{\ensuremath{\Box}}

\makeatletter
\newtheoremstyle{nonumberplainnobrackets}%
{\item[\theorem@headerfont\hskip\labelsep ##1\theorem@separator]}%
{\item[\theorem@headerfont\hskip \labelsep ##1\ ##3\theorem@separator]}
\makeatother
\theoremstyle{plain}
\theoremsymbol{\ensuremath{\blacksquare}} 
\newtheorem{claimproof}{Proof of Claim}

\usepackage[pagebackref=true,breaklinks=true,letterpaper=true,colorlinks,bookmarks=false]{hyperref}

\threedvfinalcopy 

\def\threedvPaperID{59} 

\ifthreedvfinal\fi
\begin{document}

\title{PointFlowNet: Learning Representations for\\Rigid Motion Estimation from Point Clouds}

\author{Aseem Behl \qquad
    Despoina Paschalidou \qquad
 	Simon Donné \qquad
	Andreas Geiger\\
	Autonomous Vision Group, MPI for Intelligent Systems and University of Tübingen\\
	{\tt\small \{aseem.behl,despoina.paschalidou,simon.donne,andreas.geiger\}@tue.mpg.de}
}

\maketitle

\begin{abstract}
Despite significant progress in image-based 3D scene flow estimation, the performance of such approaches has not yet reached the fidelity required by many applications. Simultaneously, these applications are often not restricted to image-based estimation: laser scanners provide a popular alternative to traditional cameras, for example in the context of self-driving cars, as they directly yield a 3D point cloud. In this paper, we propose to estimate 3D motion from such unstructured point clouds using a deep neural network. In a single forward pass, our model jointly predicts 3D scene flow as well as the 3D bounding box and rigid body motion of objects in the scene. While the prospect of estimating 3D scene flow from unstructured point clouds is  promising, it is also a challenging task. We show that the traditional global representation of rigid body motion prohibits inference by CNNs, and propose a translation equivariant representation to circumvent this problem. For training our deep network, a large dataset is required. Because of this, we augment real scans from KITTI with virtual objects, realistically modeling occlusions and simulating sensor noise. A thorough comparison with classic and learning-based techniques highlights the robustness of the proposed approach.
\end{abstract}

\vspace{-0.5cm}
\section{Introduction}
\label{sec:introduction}

For intelligent systems such as self-driving cars, the precise understanding of their surroundings is key.
Notably, in order to make predictions and decisions about the future, tasks like navigation and planning require knowledge about the 3D geometry of the environment as well as about the 3D motion of other agents in the scene.

\begin{figure*}[t!]
    \centering
    
	\setlength\tabcolsep{1pt}
	\def\arraystretch{1}
	\def\imgw{0.42\textwidth}
	
    \hspace*{\fill}
    \begin{minipage}[c]{0.03\linewidth}
    \rotatebox{90}{ISF \cite{Behl2017ICCV}}
    \end{minipage}
    \begin{minipage}[c]{\imgw}
    \includegraphics[width=\linewidth]{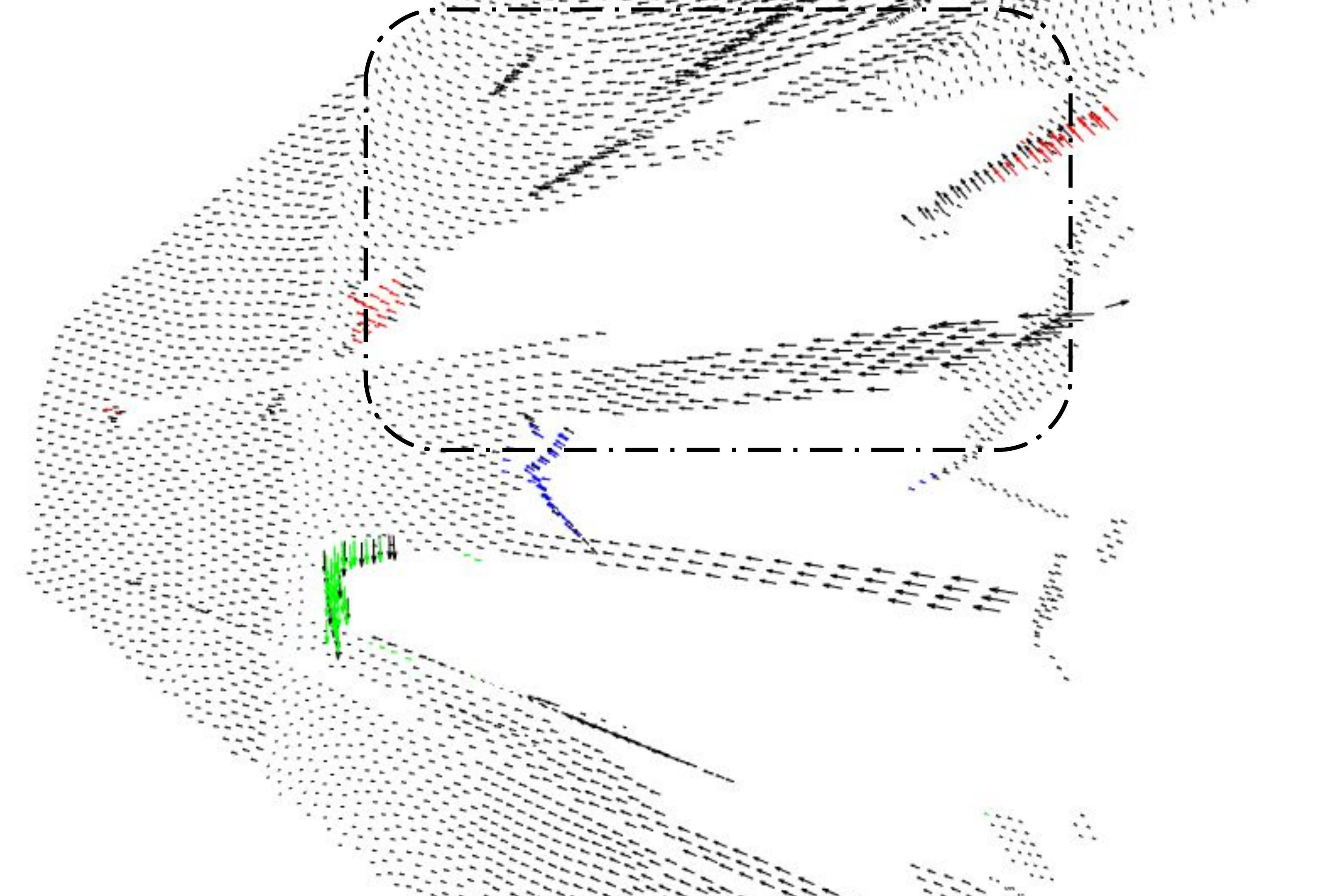}%
    \end{minipage}
    \hspace*{\fill}
    \begin{minipage}[c]{\imgw}
    \includegraphics[width=\linewidth]{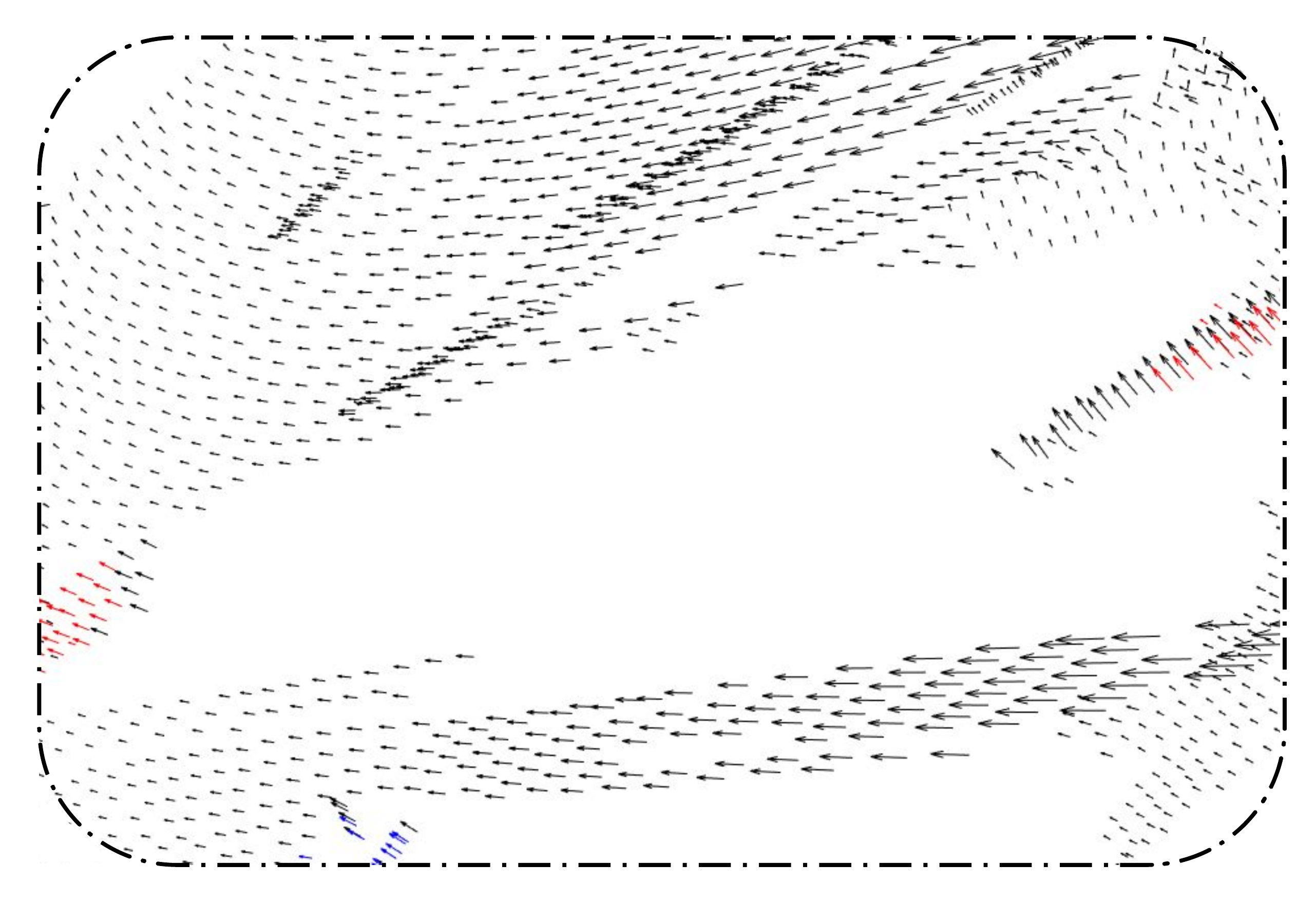}%
    \end{minipage}
    \hspace*{\fill}
    
    \hspace*{\fill}
    \begin{minipage}[c]{0.03\linewidth}
    \rotatebox{90}{PointFlowNet}
    \end{minipage}
    \begin{minipage}[c]{\imgw}
    \includegraphics[width=\linewidth]{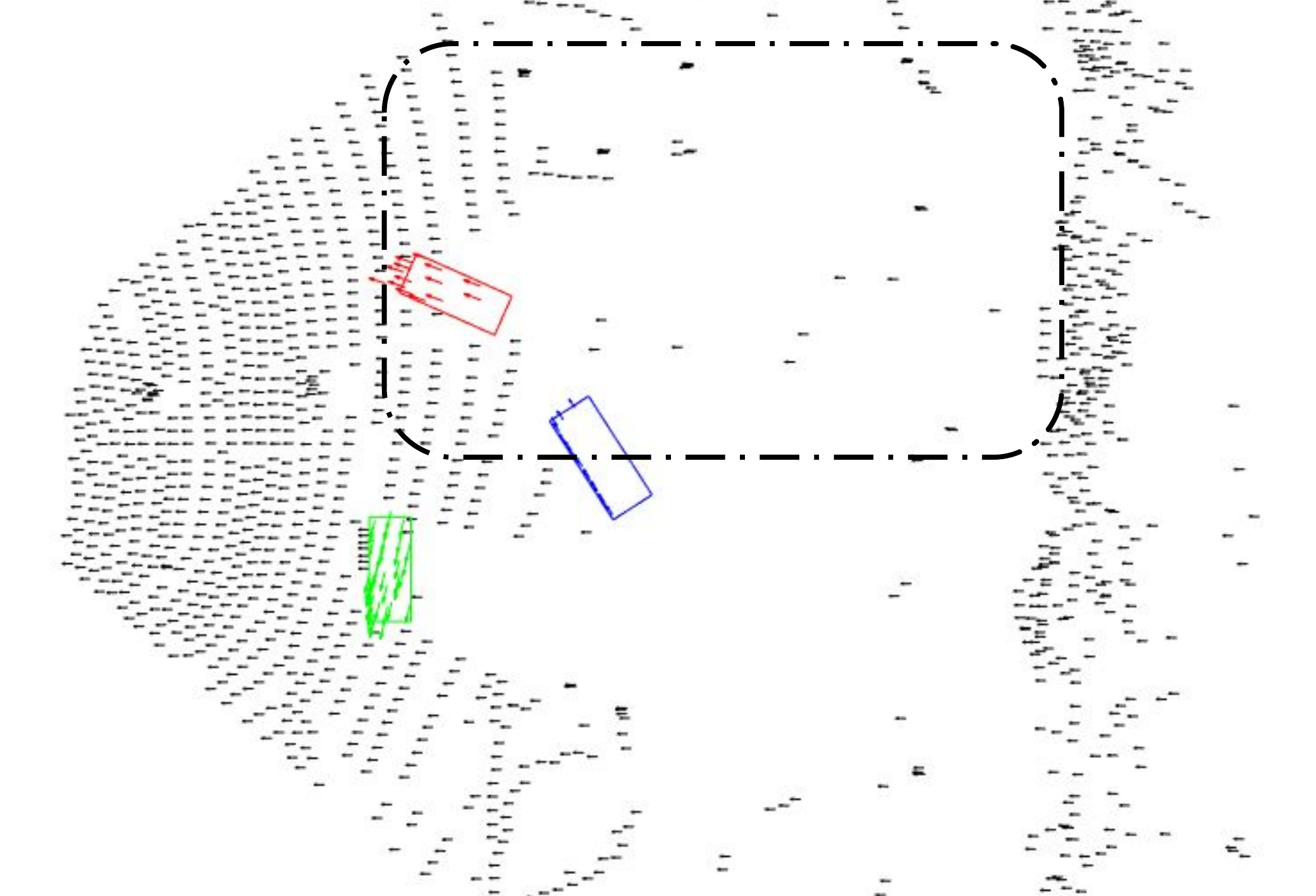}%
    \end{minipage}
    \hspace*{\fill}
    \begin{minipage}[c]{\imgw}
    \includegraphics[width=\linewidth]{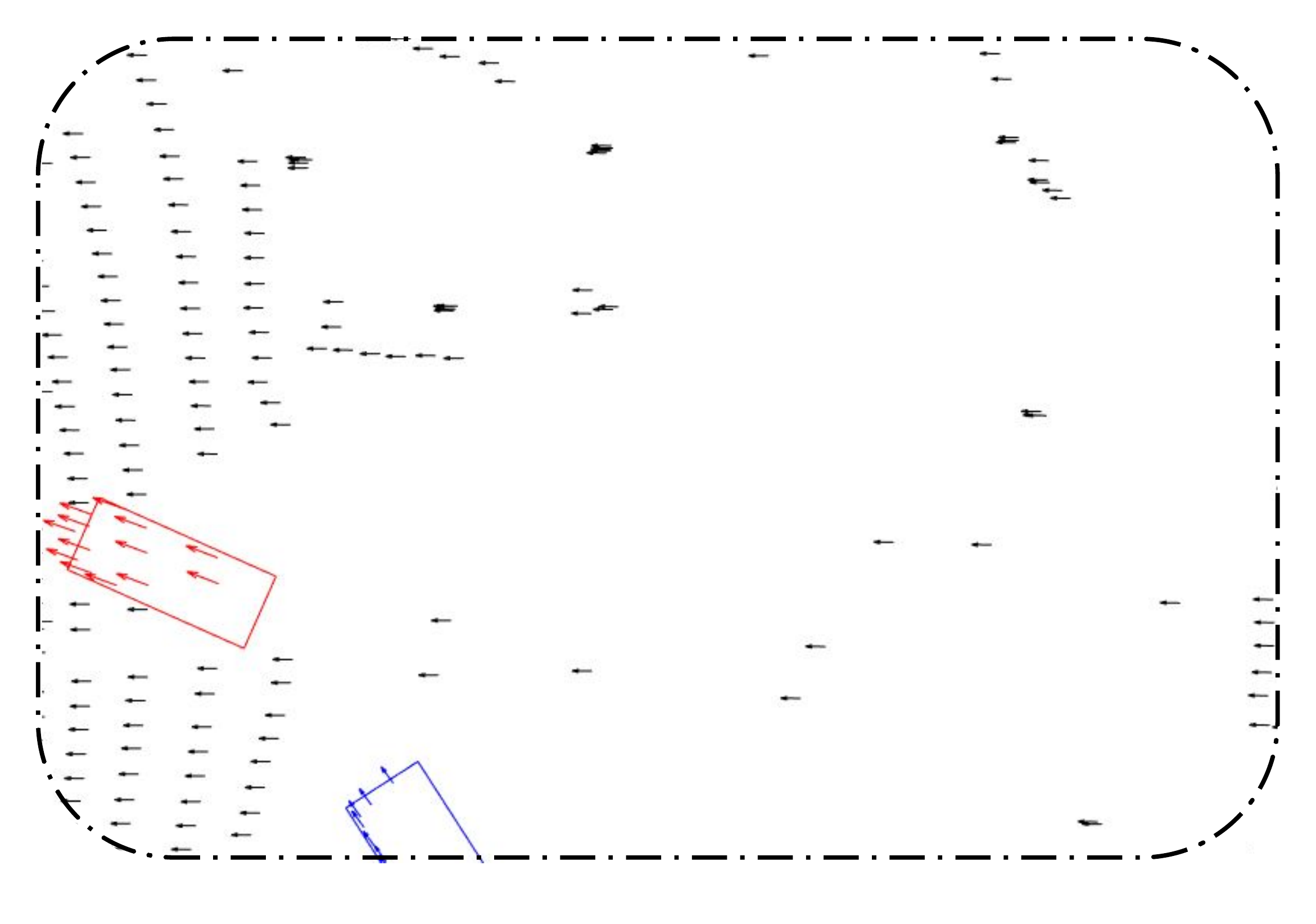}%
    \end{minipage}
    \hspace*{\fill}
    
	\caption{
		{\bf Motivation.}
		To motivate the use of LIDAR sensors in the context of autonomous driving, we provide a qualitative comparison of the state-of-the-art image-based scene flow method ISF \cite{Behl2017ICCV} (top) to our LIDAR-based PointFlowNet (bottom) using a scene from the KITTI 2015 dataset \cite{Menze2015CVPR}. The left column shows the output of the two methods. The right column shows a zoomed-in version of the inlet. While the image-based result suffers from the ``curse of two-view geometry'' (with noisy geometry, and non-uniform background movement), our LIDAR-based approach is also accurate in distant regions. Moreover, ISF relies on instance segmentation in the image space for detecting objects: depth estimation errors at the boundaries lead to objects being split into two 3D clusters (e.g., the red car). For clarity, we visualize only a subset of the points.
	}
	\label{fig:motivation}
	\vspace{-0.5cm}
\end{figure*}

3D scene flow is the most generic representation of this 3D motion; it associates a velocity vector with 3D motion to each measured point.
Traditionally, 3D scene flow is estimated based on two consecutive image pairs of a calibrated stereo rig \cite{Vedula1999CVPR,Vedula2005PAMI,Huguet2007ICCV}.
While the accuracy of scene flow methods has greatly improved over the last decade~\cite{Menze2015CVPR}, image-based scene flow methods have rarely made it into robotics applications.
The reasons for this are two-fold.
First of all, most leading techniques take several minutes or hours to predict 3D scene flow.
Secondly, stereo-based scene flow methods suffer from a fundamental flaw, the ``curse of two-view geometry'': it can be shown that the depth error grows quadratically with the distance to the observer~\cite{Lenz2011IV}.
This causes problems for the baselines and object depths often found in self-driving cars, as illustrated in \figref{fig:motivation} (top).

Consequently, most modern self-driving car platforms rely on LIDAR technology for 3D geometry perception.
In contrast to cameras, laser scanners provide a 360 degree field of view with just one sensor, are generally unaffected by lighting conditions, and do not suffer from the quadratic error behavior of stereo cameras.
However, while LIDAR provides accurate 3D point cloud measurements, estimating the motion between two such scans is a non-trivial task.
Because of the sparse and non-uniform nature of the point clouds, as well as the missing appearance information, the data association problem is complicated.
Moreover, characteristic patterns produced by the scanner, such as the circular rings in \figref{fig:motivation} (bottom), move with the observer and can easily mislead local correspondence estimation algorithms.

To address these challenges, we propose {\it PointFlowNet}, a generic model for learning 3D scene flow from pairs of unstructured 3D point clouds.
Our main contributions are:
\begin{itemize}
	\item We present an end-to-end trainable model for joint 3D scene flow and rigid motion prediction and 3D object detection from unstructured LIDAR data, as captured from a (self-driving) car.
	\item We show that a global representation is not suitable for rigid motion prediction, and propose a local translation-equivariant representation to mitigate this problem.
	\item We augment the KITTI dataset with virtual cars, taking into account occlusions and simulating sensor noise, to provide more (realistic) training data.
	\item We demonstrate that our approach compares favorably to the state-of-the-art.
\end{itemize}
We will make the code and dataset available. 
\section{Related Work}
\label{sec:related}

In the following discussion, we first group related methods based on their expected input; we finish this section with a discussion of learning-based solutions.

\boldparagraph{Scene Flow from Image Sequences}
The most common approach to 3D scene flow estimation is to recover correspondences between two calibrated stereo image pairs.
Early approaches solve the problem using coarse-to-fine variational optimization \cite{Vedula1999CVPR,Vedula2005PAMI,Basha2013IJCV,Huguet2007ICCV,Valgaerts2010ECCV,Wedel2011IJCV,Vogel2011ICCV}.
As coarse-to-fine optimization often performs poorly in the presence of large displacements, slanted-plane models which decompose the scene into a collection of rigidly moving 3D patches have been proposed \cite{Vogel2015IJCV,Menze2015CVPR,Menze2015ISA,Lv2016ECCV}. The benefit of incorporating semantics has been demonstrated in \cite{Behl2017ICCV}.
While the state-of-the-art in image-based scene flow estimation has advanced significantly, its accuracy is inherently limited by the geometric properties of two-view geometry as previously mentioned and illustrated in Figure~\ref{fig:motivation}.

\boldparagraph{Scene Flow from RGB-D Sequences}
When per-pixel depth information is available, two consecutive RGB-D frames are sufficient for estimating 3D scene flow.
Initially, the image-based variational scene flow approach was extended to \mbox{RGB-D} inputs~\cite{Wedel2008ECCV,Herbst2013ICRA,Quiroga2014ECCV}.
Franke \etal~\cite{Franke2005DAGM} instead proposed to track KLT feature correspondences using a set of Kalman filters.
Exploiting PatchMatch optimization on spherical 3D patches, 
Hornacek \etal~\cite{Hornacek2014CVPR} recover a dense field of 3D rigid body motions.
However, while structured light scanning techniques (\eg, Kinect) are able to capture indoor environments, dense RGB-D sequences are hard to acquire in outdoor scenarios like ours. Furthermore, structured light sensors suffer from the same depth error characteristics as stereo techniques.

\boldparagraph{Scene Flow from 3D Point Clouds}
In the robotics community, motion estimation from 3D point clouds has so far been addressed primarily with classical techniques.
Several works \cite{Danescu2011TITS,Ushani2017ICRA,Tanzmeister2014ICRA} extend occupancy maps to dynamic scenes by representing moving objects via particles which are updated using particle filters \cite{Danescu2011TITS,Tanzmeister2014ICRA} or EM \cite{Ushani2017ICRA}.
Others tackle the problem as 3D detection and tracking using mean shift \cite{Asvadi2016ITSC}, RANSAC \cite{Dewan2016ICRA}, ICP \cite{Moosmann2013ICRA}, CRFs \cite{Ven2010ICRA} or Bayesian networks \cite{Held2016IJRR}.
In contrast, Dewan \etal \cite{Dewan2016IROS} propose a 3D scene flow approach where local SHOT descriptors \cite{Tombari2010ECCV} are associated via a CRF that incorporates local smoothness and rigidity assumptions.
While impressive results have been achieved, all the aforementioned approaches require significant engineering and manual model specification.
In addition, local shape representations such as SHOT \cite{Tombari2010ECCV} often fail in the presence of noisy or ambiguous inputs.
In contrast, we address the scene flow problem using a generic end-to-end trainable model which is able to learn local and global statistical relationships directly from data.
Accordingly, our experiments show that our model compares favorably to the aforementioned classical approaches.

\boldparagraph{Learning-based Solutions}
While several learning-based approaches for stereo \cite{Kendall2017ICCV,Zbontar2016JMLR,Liang2017ARXIV} and optical flow \cite{Sun2018CVPR,Dosovitskiy2015ICCV,Ilg2017CVPR} have been proposed in literature, there is little prior work on learning scene flow estimation.
A notable exception is SceneFlowNet \cite{Mayer2016CVPR}, which concatenates features from FlowNet \cite{Dosovitskiy2015ICCV} and DispNet \cite{Mayer2016CVPR} for {\it image-based} scene flow estimation.
In contrast, this paper proposes a novel end-to-end trainable approach for scene flow estimation from unstructured {\it 3D point clouds}. 
More recently, Wang \etal \cite{Wang2018CVPRb} proposed a novel continuous convolution operation and applied it to 3D segmentation and scene flow. However, they do not consider rigid motion estimation which is the main focus of this work.

\section{Method}
\label{sec:method}

\begin{figure*}[ht!]
    \centering
	\includegraphics[width=0.96\linewidth]{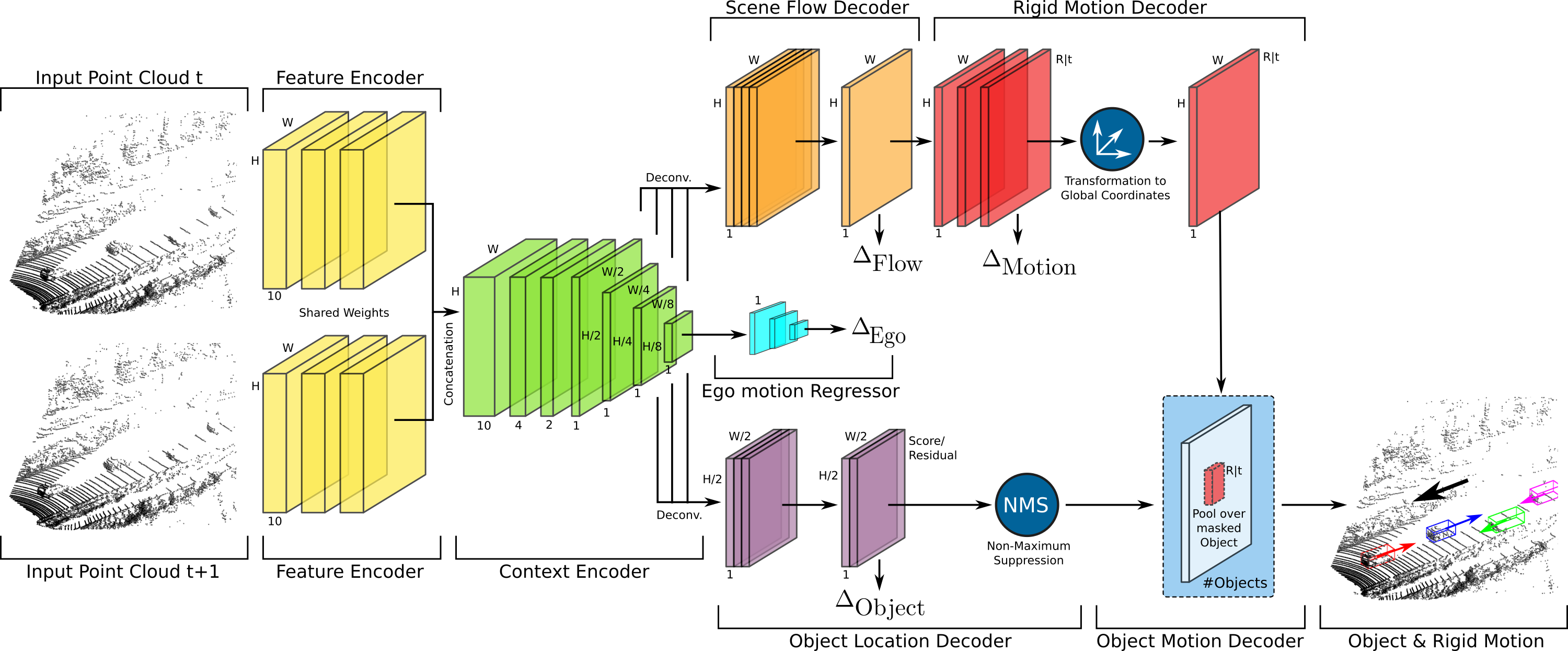}\\
        \caption{
        {\bf Network Architecture.}
        The \boldyellow{feature encoder} takes a raw LIDAR point cloud as input, groups the points into $W\times H\times 10$ voxels, and outputs 128D feature maps (for clarity, the size of the feature maps is not shown in the figure) which are concatenated and passed to the \boldgreen{context encoder}.
        The context encoder learns a global representation by interleaving convolution with strided convolution layers and ``flattening'' the third dimension (height above ground), \ie, we assume that 3D objects cannot be located on top of each other and that 3D scene points that project to the same location in the ground plane undergo the same 3D motion.
        Feature maps at different resolutions are upsampled, stacked and fed into the decoding branches.
        3D scene flow is computed for every input voxel in the \boldorange{scene flow decoder} and the result is passed to the \boldred{rigid motion decoder}, which infers a rigid body transformation for every point.
        In parallel, the \boldcyan{ego-motion regressor}, further downsamples the feature map by interleaving convolutional layers with strided convolutional layers and a fully connected layer at the end to regress rigid motion for the ego vehicle.
        In addition, the \boldpurple{object decoder} predicts the location and size (\ie, 3D bounding box) of objects in the scene.
        Finally, the \boldblue{object motion decoder} takes the point-wise rigid body motions as input and predicts the object rigid motions by pooling the rigid motion field over the detected 3D objects.
        } 
	\label{fig:architecture}
	\vspace{-0.5cm}
\end{figure*}

We start by formally defining our problem.
Let $\bP_{t}\in\nR^{N\times 3}$ and $\bP_{t+1}\in\nR^{M\times 3}$ denote the input 3D point clouds at frames $t$ and $t+1$, respectively. Our goal is to estimate
\begin{itemize}
	\item the 3D scene flow $\bv_i\in\nR^3$ and the 3D rigid 
	motion $\bR_i\in\nR^{3\times 3}$, $\bt_i\in\nR^{3}$  at each of the $N$ points in the reference point cloud at frame $t$, and
	\item the location, orientation, size and rigid motion of every moving object in the scene (in our experiments, we focus solely on cars).
\end{itemize}
The overall network architecture of our approach is illustrated in Figure~\ref{fig:architecture}.
The network comprises four main components: (1) feature encoding layers, (2) scene flow estimation, ego-motion estimation and 3D object detection layers, (3) rigid motion estimation layers and (4) object motion decoder. In the following, we provide a detailed description for each of these components as well as the loss functions.

\subsection{Feature Encoder}

The feature encoding layers take a raw point cloud as input, partition the space into voxels, and describe each voxel with a feature vector.
The simplest form of aggregation is binarization, where any voxel containing at least one point is set to $1$ and all others are zero.
However, better results can be achieved by aggregating high-order statistics over the voxel~
\cite{Qi2017CVPR,Qi2017NIPS,Zhou2018CVPR,Su2018CVPR,Purkait2017ARXIV,Chen2017CVPR}.
In this paper, we leverage the feature encoding recently proposed by Zhou \etal \cite{Zhou2018CVPR}, which has demonstrated state-of-the-art results for 3D object detection from point clouds.

We briefly summarize this encoding, but refer the reader to \cite{Zhou2018CVPR} for more details.
We subdivide the 3D space of each input point cloud into equally spaced voxels
and group points according to the voxel they reside in.
To reduce bias \wrt LIDAR point density, a fixed number of T points is randomly sampled for all voxels containing more than T points.
Each voxel is processed with a stack of Voxel Feature Encoding (VFE) layers to capture local and global geometric properties of its contained points.
As more than $90\%$ of the voxels in LIDAR scans tend to be empty, we only process non-empty voxels and store the results in a sparse 4D tensor.


We remark that alternative representations, \eg, those that directly encode the raw point cloud \cite{Wang2018CVPRb,Groh2018ACCV}, could be a viable alternative to voxel representations. However, as the representation is not the main focus of this paper, we will leave such an investigation to future work.


\subsection{3D Detection, Ego-motion and 3D Scene Flow}

As objects in a street scene are restricted to the ground plane, we only estimate objects and motions on this plane: we assume that 3D objects cannot be located on top of each other and that 3D scene points directly above each other undergo the same 3D motion. 
This is a valid assumption for our autonomous driving scenario, and greatly improves memory efficiency.
Following \cite{Zhou2018CVPR}, we vertically downsample the voxel feature map to size $1$ by using three 3D convolutions with vertical stride $2$.

The resulting 3D feature map is reshaped by stacking the remaining height slices as feature maps to yield a 2D feature map.
The first layer of each block downsamples the feature map via a convolution with stride $2$, followed by a series of convolution layers with stride $1$.
Each convolution layer is followed by Batch Normalization and a ReLU.

Next, the network splits up in three branches for respectively ego-motion estimation, 3D object detection and 3D scene flow estimation.
As there is only one observer, the ego-motion branch further downsamples the feature map by interleaving convolutional layers with strided convolutional layers and finally using a fully connected layer to regress a 3D ego-motion (movement in the ground-plane and rotation around the vertical).
For the other two tasks, we upsample the output of the various blocks using up-convolutions: to half the original resolution for 3D object detection, and to the full resolution for 3D scene flow estimation.
The resulting features are stacked and mapped to the training targets with one 2D convolutional layer each.
We regress a 3D vector per voxel for the scene flow, and follow \cite{Zhou2018CVPR} for the object detections: regressing likelihoods for a set of proposal bounding boxes and regressing the residuals (translation, rotation and size) between the positive proposal boxes and corresponding ground truth boxes. 
A proposal bounding box is called positive if it has the highest Intersection over Union (IoU, in the ground plane) with a ground truth detection, or if its IoU with any ground truth box is larger than 0.6, as in~\cite{Zhou2018CVPR}.

\subsection{Rigid Motion Decoder}

We now wish to infer per-pixel and per-object rigid body motions from the previously estimated 3D scene flow.
For a single point in isolation, there are infinitely many rigid body motions that explain a given 3D scene flow: this ambiguity can be resolved by considering the local neighborhood.

It is unfortunately impossible to use a convolutional neural network to regress rigid body motions that are represented in global world coordinates, as the conversion between scene flow and global rigid body motion depends on the location in the scene: while convolutional layers are translation equivariant, the mapping to be learned is not.
Identical regions of flow lead to different global rigid body motions, depending on the location in the volume, and a fully convolutional network cannot model this.
In the following, we first prove that the rigid motion in the world coordinate system is not translation equivariant.
Subsequently, we introduce our proposed rigid motion representation in local coordinates and show it to be translation equivariant and therefore amenable to fully convolutional inference.

\begin{figure*}[t!]
	\centering
	\hspace*{\fill}
	\begin{subfigure}[b]{0.47\linewidth}
	\includegraphics[width=0.9\linewidth]{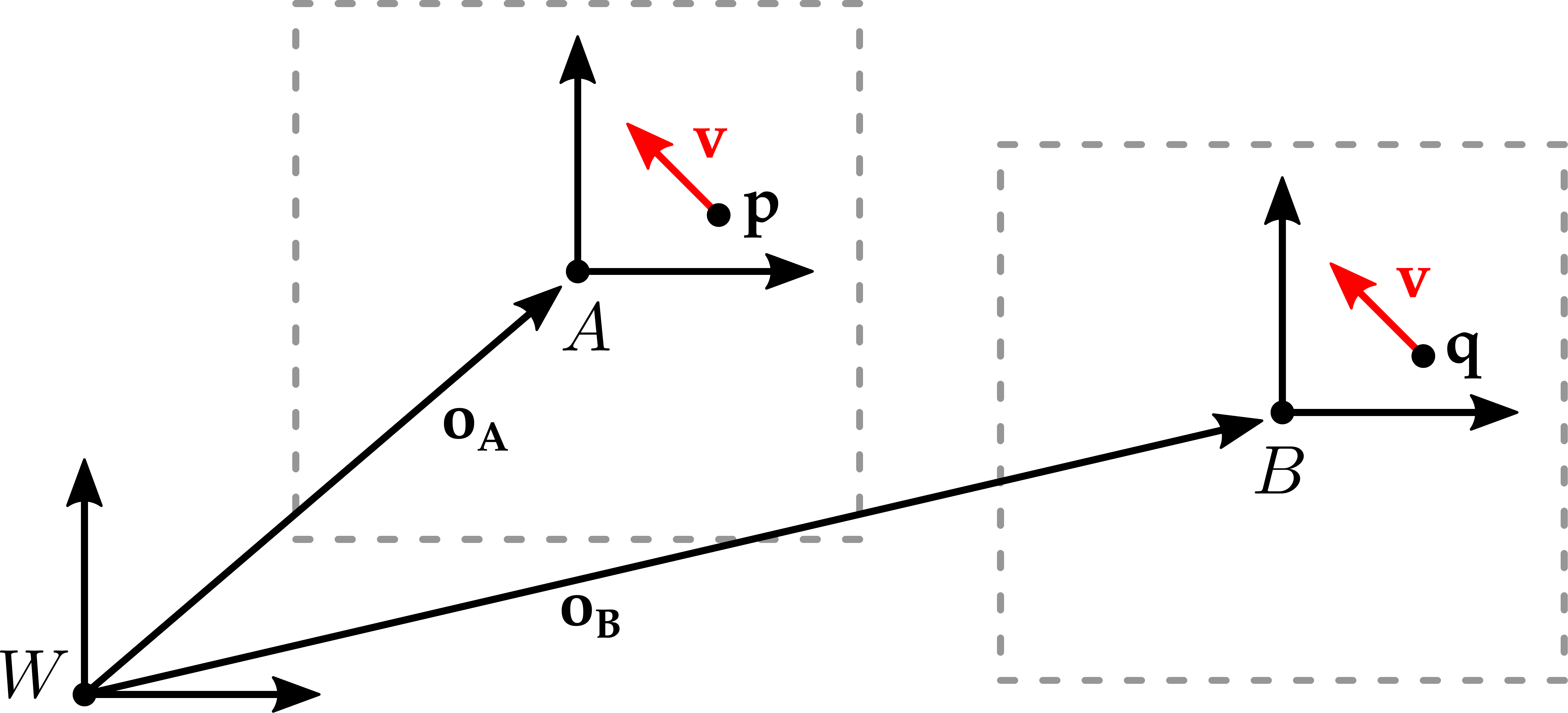}
	\vspace{0.2cm}
	\caption{Local (A,B) and World Coordinate System (W)}
	\label{fig:equivariance_illustration}
	\end{subfigure}
	\hspace*{\fill}
	\begin{subfigure}[b]{0.47\linewidth}
	\includegraphics[width=\linewidth]{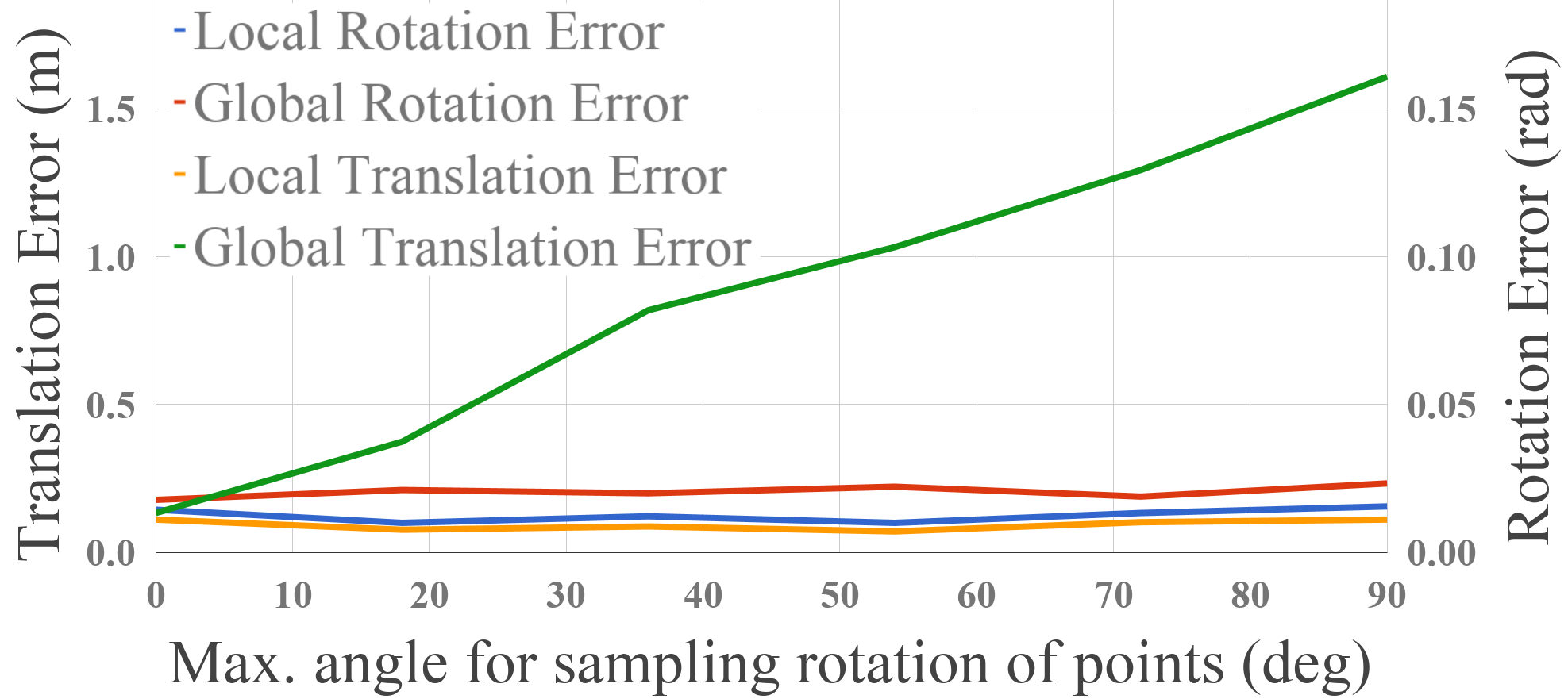}
	\caption{Quantitative Comparison}\label{fig:equivariance_results}
	\end{subfigure}
	\hspace*{\fill}
	\caption{{\bf Rigid Motion Estimation.} In (\subref{fig:equivariance_illustration}), indices $A$ and $B$ denote the coordinate system of points $\bp$ and $\bq$ at origin $\vec{o}_A$ and $\vec{o}_B$, respectively. The same scene flow $\bv$ can locally be explained with the same rigid body motion $(\bR_L,\bt_L)$, but requires different translations $\bt^\vec{p}_{W}\neq\bt^\vec{q}_{W}$ in the global coordinate system.
	A simple example (\subref{fig:equivariance_results}) provides empirical evidence that translation cannot be learned in global coordinates with a CNN. Using global coordinates, the translation error increases significantly with the magnitude of rotation (\textcolor{darkgreen}{green}). There is no such increase in error when using local coordinates (\textcolor[rgb]{1,0.5,0.3}{orange}).}
	\label{fig:equivariance}
	\vspace{-0.3cm}
\end{figure*}

Let us assume a point $\vec{p}$ in world coordinate system $W$
and let $A$ denote a local coordinate system with origin $\bo_A$ as illustrated in \figref{fig:equivariance_illustration}.
A scene flow vector $\bv$ is explained by rigid body motion $(\bR_{A},\bt_{A})$, represented in local coordinate system $A$ with origin $\bo_A$, if and only if:
\begin{equation} 
\vec{v} = \left[\bR_{A} \left(\vec{p} - \vec{o}_A\right)  + \vec{t}_{A})\right] - \left(\vec{p} - \vec{o}_A\right)
\end{equation}
Now assume a second world location $\vec{q}$, also with scene flow $\bv$ as in \figref{fig:equivariance_illustration}.
Let $B$ denote a second local coordinate system with origin $\bo_B$ such that $\bp$ and $\bq$ have the same local coordinates in their respective coordinate system, \ie, $\vec{p}-\vec{o}_A = \vec{q} - \vec{o}_B$.
We now prove the following two claims: 
\begin{enumerate}
	\item There exists no rigid body motion $\bR_{W},\bt_{W}$ represented in world coordinate system $W$ that explains the scene flow $\vec{v}$ for both $\vec{p}$ and $\vec{q}$, unless $\bR_{W} = \bI$.
	\item Any rigid body motion $(\bR_{A},\bt_{A})$ explaining scene flow $\vec{v}$ for $\vec{p}$ in system $A$ also does so for $\vec{q}$ in system $B$.
\end{enumerate}
%
%
Towards this goal, we introduce the notation $(\bR^\vec{p}_{W},\bt^\vec{p}_{W})$ to indicate rigid motion in world coordinates $W$ induced by $\bv_\bp$.
\begin{claim}
\begin{equation}
\begin{aligned}
    \forall \vec{p}, \vec{q} \in \mathbb{R}^3, \vec{p} - \vec{o}_A = &~ \vec{q} - \vec{o}_B, \vec{o}_A \neq \vec{o}_B:\\
    \vec{v}_{\vec{p}} = \vec{v}_{\vec{q}} \implies &\bR^\vec{p}_{W} \neq \bR^\vec{q}_{W} \quad \textrm{ or } \\&\bt^\vec{p}_{W} \neq \bt^\vec{q}_{W} \quad \textrm{ or } \\&\bR^\vec{p}_{W} = \bR^\vec{p}_{W} = \bI
\end{aligned}
\end{equation}
\end{claim}
\begin{claimproof}
    From $\vec{v}_{\vec{p}} = \vec{v}_{\vec{q}}$ we get
        \begin{equation*}
    \resizebox{\linewidth}{!}{$
        \begin{aligned}
            \bR^\vec{p}_{W} \vec{p} + \bt^\vec{p}_{W} - \vec{p} &= \bR^\vec{q}_{W} \vec{q} + \bt^\vec{q}_{W} - \vec{q} \\
            \bR^\vec{p}_{W} \vec{p} + \bt^\vec{p}_{W} &= \bR^\vec{q}_{W} \left(\vec{p}-{\Delta}\vec{o}\right) + \bt^\vec{q}_{W} + {\Delta}\vec{o}\\
            \left(\bR^\vec{p}_{W} - \bR^\vec{q}_{W}\right)\vec{p} &= \left(\bI - \bR^\vec{q}_{W} \right){\Delta}\vec{o} + \left(\bt^\vec{q}_{W} - \bt^\vec{p}_{W}\right)
        \end{aligned}
    $}
        \end{equation*}
        where ${\Delta}\vec{o} = \vec{o}_A - \vec{o}_B$.
        Now, we assume that $\bR^\vec{p}_{W} = \bR^\vec{q}_{W}$ and that $\bt^\vec{p}_{W} = \bt^\vec{q}_{W}$ (in all other cases the claim is already fulfilled).
        In this case, we have ${\Delta}\vec{o} = \bR^\vec{p}_{W} {\Delta}\vec{o}$.
        However, any rotation matrix representing a non-zero rotation has no real eigenvectors.
        Hence, as $\vec{o}_A \neq \vec{o}_B$, this equality can only be fulfilled if $\bR^\vec{p}_{W}$ is the identity matrix.
        \hfill$\blacksquare$
\end{claimproof}

\begin{claim}
\begin{equation}
\begin{aligned}
    \forall \vec{p}, \vec{q} \in \mathbb{R}^3, &~\vec{p} - \vec{o}_A = \vec{q} - \vec{o}_B, \vec{o}_A \neq \vec{o}_B:\\
    &\vec{v} = \bR\left(\vec{p} - \vec{o}_A\right) + \bt + \left(\vec{p}-\vec{o}_A\right)\\\implies &\vec{v} = \bR\left(\vec{q} - \vec{o}_B\right) + \bt + \left(\vec{q} - \vec{o}_B\right)
\end{aligned}
\end{equation}
\end{claim}
\begin{claimproof}
    Trivially from $\vec{p} - \vec{o}_A = \vec{q} - \vec{o}_B$. \hfill$\blacksquare$
\end{claimproof}

The first proof shows the non-stationarity of rigid body motions represented in global coordinates, while the second proof shows that the rigid motion represented in local coordinates is stationary and can therefore be learned by a translation equivariant convolutional neural network.

We provide a simple synthetic experiment in Figure~\ref{fig:equivariance} to empirically confirm this analysis.
Towards this goal, we warp a grid of $10\times10$ points by random rigid motions, and then try to infer these rigid motions from the resulting scene flow: as expected, the estimation is only successful using local coordinates.
Note that a change of reference system only affects the translation component while the rotation component remains unaffected. 
Motivated by the preceding analysis, we task our CNN to predict rigid motion in local coordinates, followed by a deterministic layer which transforms local coordinates into global coordinates as follows:
%
\begin{equation}
\begin{array}{ll}
\bR_{L} = \bR_{W} & \quad \bt_{L} = (\bR_{W}-\bI)\bo_L+\bt_{W} \\
\bR_{W} = \bR_{L} & \quad \bt_{W} = (\bI - \bR_{W})\bo_L+\bt_{L}
\end{array}
\label{eq:cs_conversion}
\end{equation}
In our case, the origin of the world coordinate system $W$ coincides with the LIDAR scanner and the origin of the local coordinate systems is located at the center of each voxel.

\subsection{Object Motion Decoder}

Finally, we combine the results of 3D object detection and rigid motion estimation into a single rigid motion for each detected object.
We first apply non-maximum-suppression (NMS) using detection threshold $\tau$, yielding a set of 3D bounding boxes.
To estimate the rigid body motion of each detection, we pool the predicted rigid body motions over the corresponding voxels (\ie, the voxels in the bounding box of the detection) by computing the median translation and rotation.
Note that this is only possible as the rigid body motions have been converted back into world coordinates.

\subsection{Loss Functions}

This section describes the loss functions used by our approach.
While it seems desirable to define a rigid motion loss directly at object level, this is complicated by the need for differentiation through the non-maximum-suppression step and the difficulty associating to ground truth objects.
Furthermore, balancing the influence of an object loss across voxels is much more complex than applying all loss functions directly at the voxel level.
We therefore use auxiliary voxel-level loss functions.
Our loss comprises four parts:
\begin{equation}
\cL = \alpha \cL_{flow} + \beta \cL_{rigmo} + \gamma \cL_{ego} + \cL_{det}
\end{equation}
Here, $\alpha, \beta, \gamma$ are positive constants for balancing the relative importance of the task specific loss functions. 
We now describe the task-specific loss functions in more detail.

\boldparagraph{Scene Flow Loss} 
The scene flow loss is defined as the average $\ell_1$ distance between the predicted scene flow and the true scene flow at every voxel
\begin{equation}
\cL_{flow} = \frac{1}{K}\sum_{j}\left\Vert\bv_j - \bv_j^*\right\Vert_{1}
\end{equation}
where $\bv_j\in\nR^3$ and $\bv_j^*\in\nR^3$ denote the regression estimate and ground truth scene flow at voxel $j$, and $K$ is the number of non-empty voxels.

\boldparagraph{Rigid Motion Loss} 
The rigid motion loss is defined as the average $\ell_1$ error between the predicted translation $\bt_{j}\in\nR^2$ and its ground truth $\bt_{j}^*\in\nR^2$ in the local coordinate system and the average $\ell_1$ error between the predicted rotation $\theta_j$ around the Z-axis and its ground truth $\theta_{j}^*$ at every voxel $j$. 
\begin{equation}
\cL_{rigmo} = \frac{1}{K}\sum_{j}\left\Vert\bt_{j} - \bt_{j}^*\right\Vert_{1} + \lambda\left\Vert\theta_j - \theta_j^*\right\Vert_{1}
\end{equation}
where $\lambda$ is a positive constant to balance the relative importance of the two terms.
The conversion from world coordinates to local coordinates is given by (see also \eqref{eq:cs_conversion})
\begin{equation}
\bR_L = \bR_W(\theta_j) ~~~~~~ \bt_L = (\bR_W(\theta_j)-\bI)\,\bp_j+\bt_W~
\end{equation}
where $\bp_j \in \nR^2$ specifies the position of voxel $j$ in the XY-plane in world coordinates and $\bR_W(\theta_j)$ is the rotation matrix corresponding to rotation $\theta_j$ around the Z-axis.

\boldparagraph{Ego-motion Loss} 
Similarly, the ego-motion loss is defined as the $\ell_1$ distance between the predicted background translation $\bt_{BG}\in\nR^2$ and its ground truth $\bt_{BG}^*\in\nR^2$ and the predicted rotation $\theta_{BG}$  and its ground truth $\theta_{BG}^*$:
\begin{equation}
\cL_{ego} = \|\bt_{BG} - \bt_{BG}^*\|_1 + \lambda\|\theta_{BG} - \theta_{BG}^*\|_1
\end{equation}

\boldparagraph{Detection Loss}
Following \cite{Zhou2018CVPR}, we define the detection loss as follows:
\begin{equation}
\begin{aligned}
\cL_{det} = &\frac{1}{M_{pos}}\sum_{k}\cL_{cls}(p^{pos}_k, 1) + \cL_{reg}(\br_k, \br^*_k	) \\
&+\frac{1}{M_{neg}}\sum_{l}	\cL_{cls}(p^{neg}_l, 0)
\end{aligned}
\end{equation}
where $p^{pos}_k$ and $p^{neg}_l$ represent the softmax output for positive proposal boxes $a^{pos}_k$ and negative proposal boxes $a^{neg}_l$, 
respectively. $\br_k \in \mathbb{R}^7$ and $\br^*_k \in \mathbb{R}^7$ denote the regression estimates and ground truth residual vectors (translation, rotation and size) for the positive proposal box $k$, respectively. 
$M_{pos}$ and $M_{neg}$ represent the number of positive and negative proposal boxes.
$\cL_{cls}$ denotes the binary cross entropy loss, while $\cL_{reg}$ represents the smooth $\ell_1$ distance function.
We refer to \cite{Zhou2018CVPR} for further details.

\section{Experimental Evaluation}
\label{sec:results}

We now evaluate the performance of our method on the KITTI object detection dataset \cite{Geiger2012CVPR} as well as an extended version, which we have augmented by simulating virtual objects in each scene.

\begingroup
\setlength{\tabcolsep}{6pt}	
\begin{table*}[h!]
\centering
		\begin{tabular}{cccccccccc}
			\toprule
			\multirow{2}{*}{Eval.} &
			\multirow{2}{*}{Training} &
			\multicolumn{3}{c}{Scene Flow (m)} &
			\multicolumn{2}{c}{Object Motion} &
			\multicolumn{2}{c}{Ego-motion}\\
			\cmidrule{3-9} 
			Dataset
			& Dataset
			& {FG}& {BG}& {All} 
			& {Rot.(rad)}& {Tr.(m)}
			& {Rot.(rad)}& {Tr.(m)}  \\
			\midrule
			K & K &  0.23 & {\bf 0.14} & {\bf 0.14} & {\bf 0.004} & 0.30 & {\bf 0.004} & {\bf 0.09} \\
			K & K+AK & {\bf 0.18} & {\bf 0.14} & {\bf 0.14} & {\bf 0.004} & {\bf 0.29} & {\bf 0.004} & {\bf 0.09}
			 \\
			\cmidrule{1-9}
			K+AK & K & 0.58 & {\bf 0.14} & 0.18  & {\bf 0.010} & 0.57 & {\bf 0.004} & 0.14 \\
			K+AK & K+AK & {\bf 0.28} & {\bf 0.14} & {\bf 0.16} & {\bf 0.011} & {\bf 0.48} & {\bf 0.004} & {\bf 0.12}
			 \\
			\bottomrule
		\end{tabular}
		\caption{{\bf Ablation Study} on our KITTI and Augmented KITTI validation datasets, abbreviated with K and AK, respectively.
		}
		\label{tab:ablation}
\end{table*}
\endgroup

\begingroup
\setlength{\tabcolsep}{6pt}	
\begin{table*}[h!]
\centering
        \begin{adjustbox}{max width=\textwidth}
		\begin{tabular}{cc*{7}c}
			\toprule
			\multirow{2}{*}{Eval.} &
			\multirow{3}{*}{Method} &
			\multicolumn{3}{c}{Scene Flow (m)} &
			\multicolumn{2}{c}{Object Motion} &
			\multicolumn{2}{c}{Ego-motion} \\
			\cmidrule{3-5} \cmidrule{6-9} 
			Dataset &
			& {FG}& {BG}& {All} 
			& {Rot.(rad)}& {Tr.(m)}
			& {Rot.(rad)}& {Tr.(m)}  \\
			\hline
			K & ICP+Det. &  0.56 & 0.43 & 0.44 & 0.22 & 6.27 & {\bf 0.004} & 0.44
			\\
			K & 3DMatch+Det. &  0.89 & 0.70 & 0.71 & 0.021 & 1.80 & {\bf 0.004} & 0.68\\
			K & FPFH+Det. &  3.83 & 4.24 & 4.21 & 0.299 & 14.23 & 0.135 & 4.27\\
			K & Dewan et al.+Det. &  0.55 & 0.41 & 0.41 & 0.008 & 0.55 & {\bf 0.006} & 0.39\\
			K & Ours & {\bf 0.29} & {\bf 0.15} & {\bf 0.16} &  {\bf 0.004} & {\bf 0.19} & {\bf 0.005} & {\bf 0.12} \\
			\hline
			
			K+AK & ICP+Det. & 0.74 & 0.48 &	0.50 & 0.226 & 6.30 & {\bf 0.005} & 0.49\\
			K+AK &3DMatch+Det. & 1.14 & 0.77 & 0.80  & 0.027 & 1.76 &	{\bf 0.004}	& 0.76\\
			K+AK & FPFH+Det. & 4.00	& 4.39 & 4.36 & 0.311 & 13.54 & 0.122 & 4.30\\	
			K+AK &Dewan et al.+Det. &  0.60 & 0.52 & 0.52 & 0.014 & 0.75 & {\bf 0.006} & 0.46\\		
			K+AK & Ours & {\bf 0.34} & {\bf 0.18} & {\bf 0.20} & {\bf 0.011} & {\bf 0.50} & {\bf 0.005} & {\bf 0.15}\\
			\bottomrule
		\end{tabular}
		\end{adjustbox}
		\caption{{\bf Comparison to Baselines} on test sets of KITTI and Augmented KITTI, abbreviated with K and AK, respectively. }
		\label{tab:results}
		\vspace{-0.5cm}
\end{table*}
\endgroup
	
\subsection{Datasets}

\boldparagraph{KITTI}
For evaluating our approach, we use 61 sequences of the training set in the KITTI object detection dataset \cite{Geiger2012CVPR}, containing a total of 20k frames.
As there is no pointcloud-based scene flow benchmark in KITTI, we  perform our experiments on the original training set.
Towards this goal, we split the original training set into 70\% train, 10\% validation, 20\% test sequences, making sure that frames from the same sequence are not used in different splits.

\begin{figure}
    \centering
    
    \hspace*{\fill}    
    \includegraphics[width=0.48\linewidth]{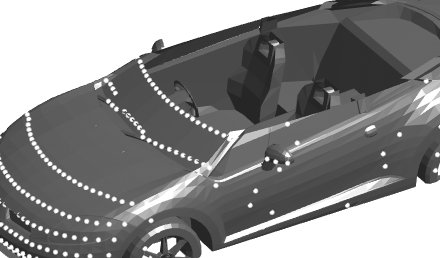}%
    \hspace*{\fill}
	\includegraphics[width=0.48\linewidth]{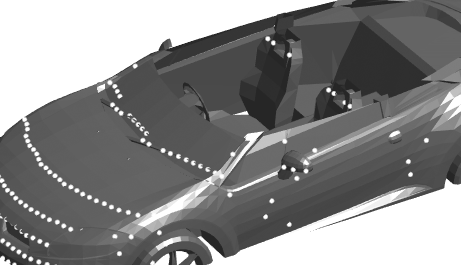}%
    \hspace*{\fill}
	\caption{{\bf Augmentation.}
	Simulating LIDAR measurements based on 3D meshes would result in measurements at transparant surfaces such as windows (\textbf{left}), wheres a real LIDAR scanner measures interior points instead.
	Our simulation replicates the behavior of LIDAR scanners by taking into account model transparency and learning the noise model from real KITTI scans  (\textbf{right}).
	}
	\label{fig:lidarSimulator}
	\vspace{-0.5cm}
\end{figure}
	
\boldparagraph{Augmented KITTI}
	However, the official KITTI object detection datasets lacks cars with a diverse range of motions.
	To generate more salient training example, we generate a realistic mixed reality LiDAR dataset exploiting a set of high quality 3D CAD models of cars~\cite{Fidler2012NIPS} by taking the characteristics of real LIDAR scans into account.
	
We discuss our workflow here.
We start by fitting the ground plane using RANSAC 3D plane fitting; this allows us to detect obstacles and hence the drivable region.
In a second step, we randomly place virtual cars in the drivable region,  and simulate a new LIDAR scan that includes these virtual cars.
Our simulator uses a noise model learned from the real KITTI scanner, and also produces missing estimates at transparent surfaces using the transparency information provided by the CAD models, as illustrated in Figure~\ref{fig:lidarSimulator}.
Additionally, we remove points in the original scan which become occluded by the augmented car by tracing a ray between each point and the LIDAR, and removing those points whose ray intersects with the car mesh.
Finally, we sample the augmented car's rigid motion using a simple approximation of the Ackermann steering geometry, place the car at the corresponding location in the next frame, and repeat the LIDAR simulation.
We generate $20$k such frames with $1$ to $3$ augmented moving cars per scene. We split the sequences into $70\%$ train, $10\%$ validation, $20\%$ test similar to our split of the original KITTI dataset.

\begin{figure*}[t!]
	\centering
	\setlength\tabcolsep{1pt}
	\def\arraystretch{1}
	\def\imgw{0.44\textwidth}
	\begin{tabular}{c c}
		\includegraphics[width=\imgw]{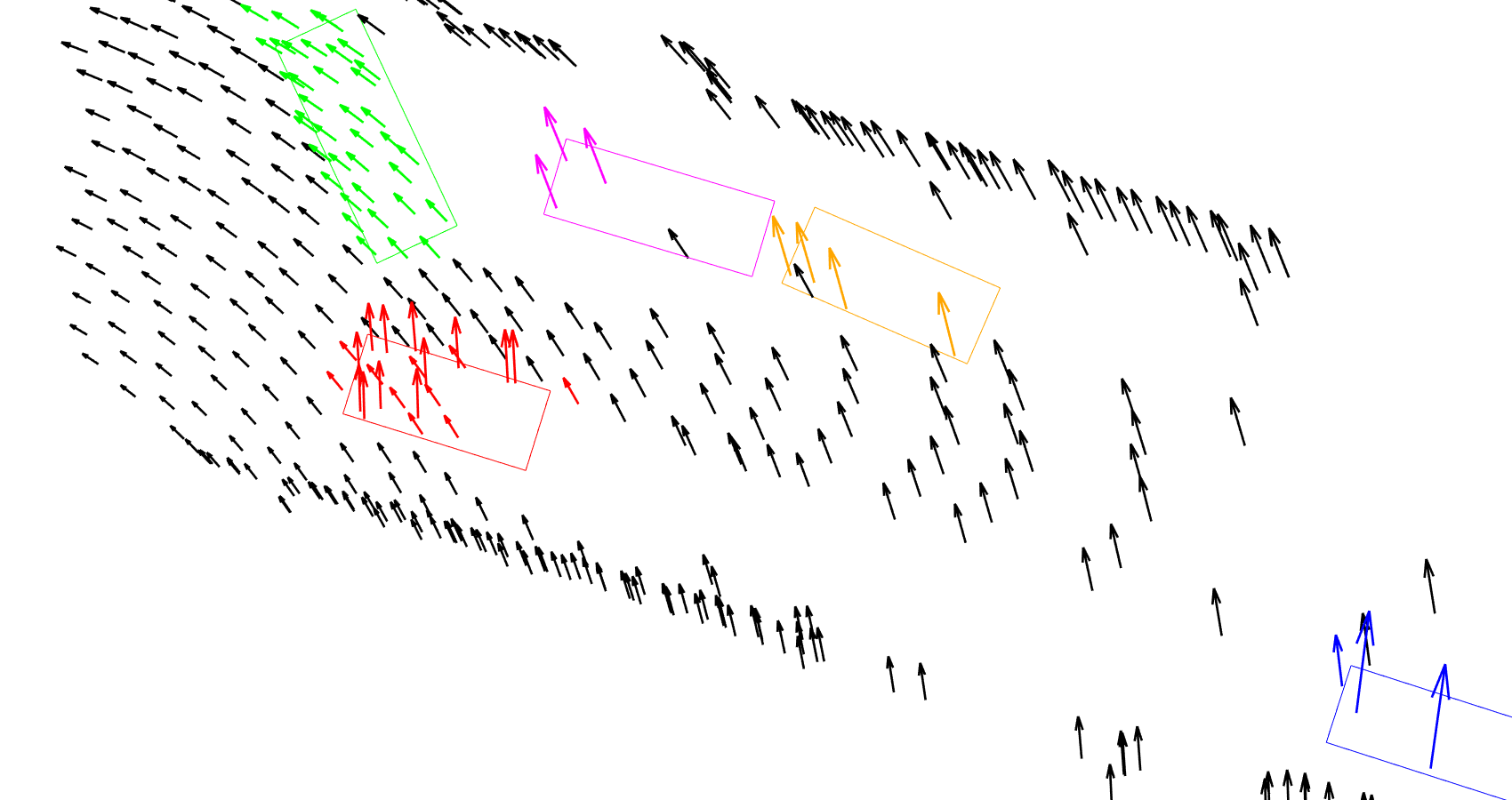}&%
		
		\includegraphics[width=\imgw]{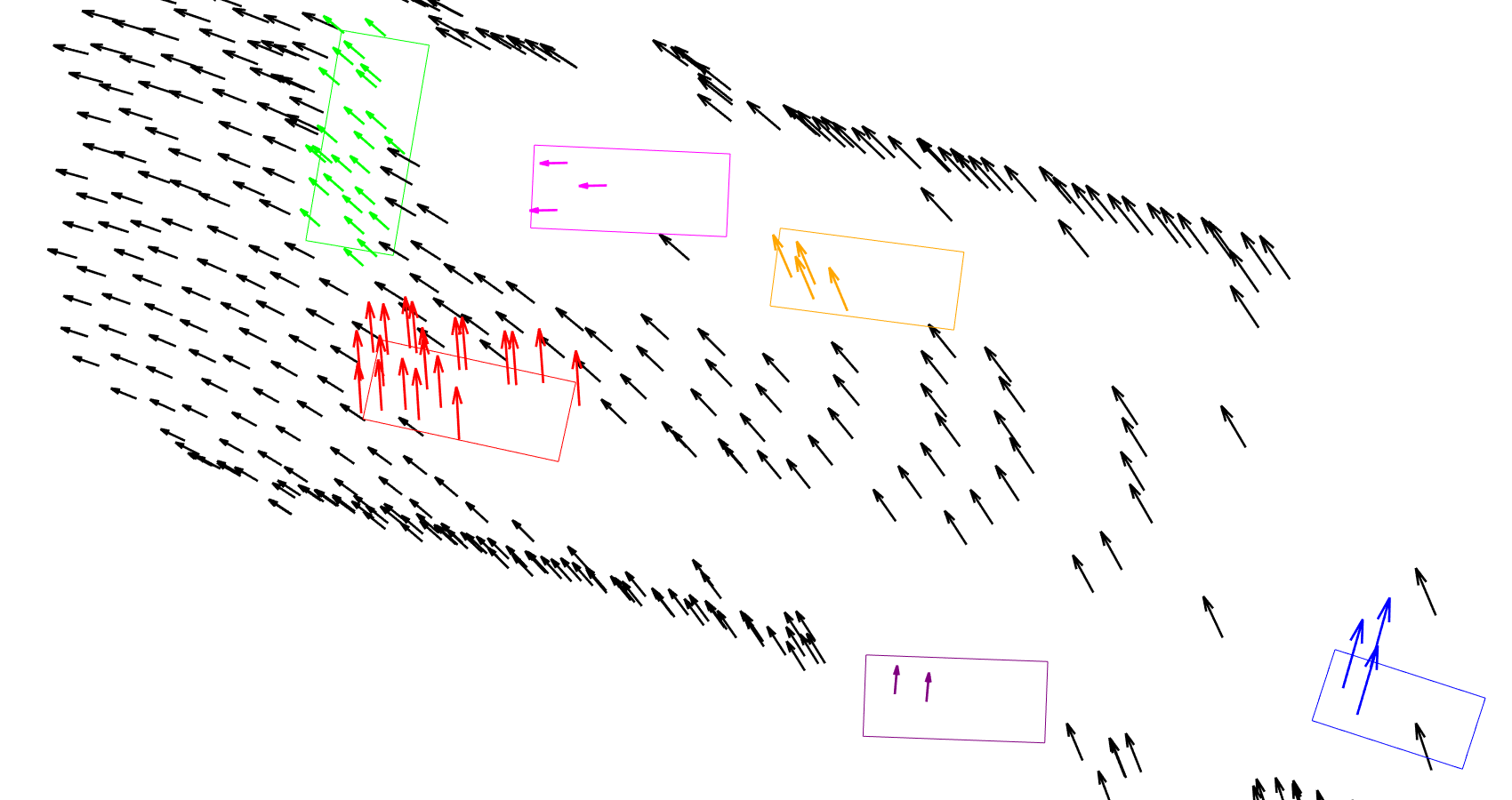}\\%
		(a) Ground Truth & (b) Our result \\
		\includegraphics[width=\imgw]{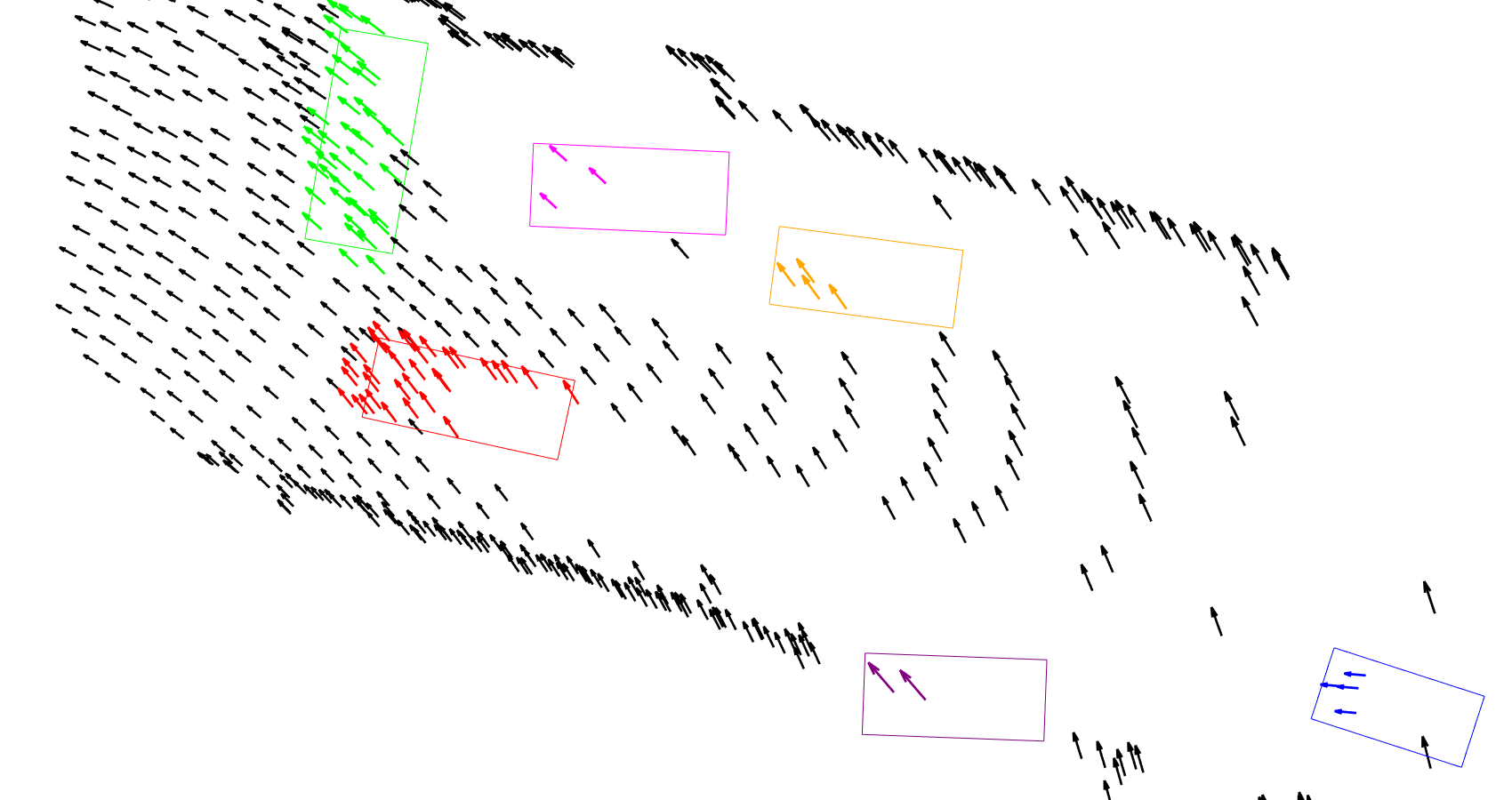}&%
		\includegraphics[width=\imgw]{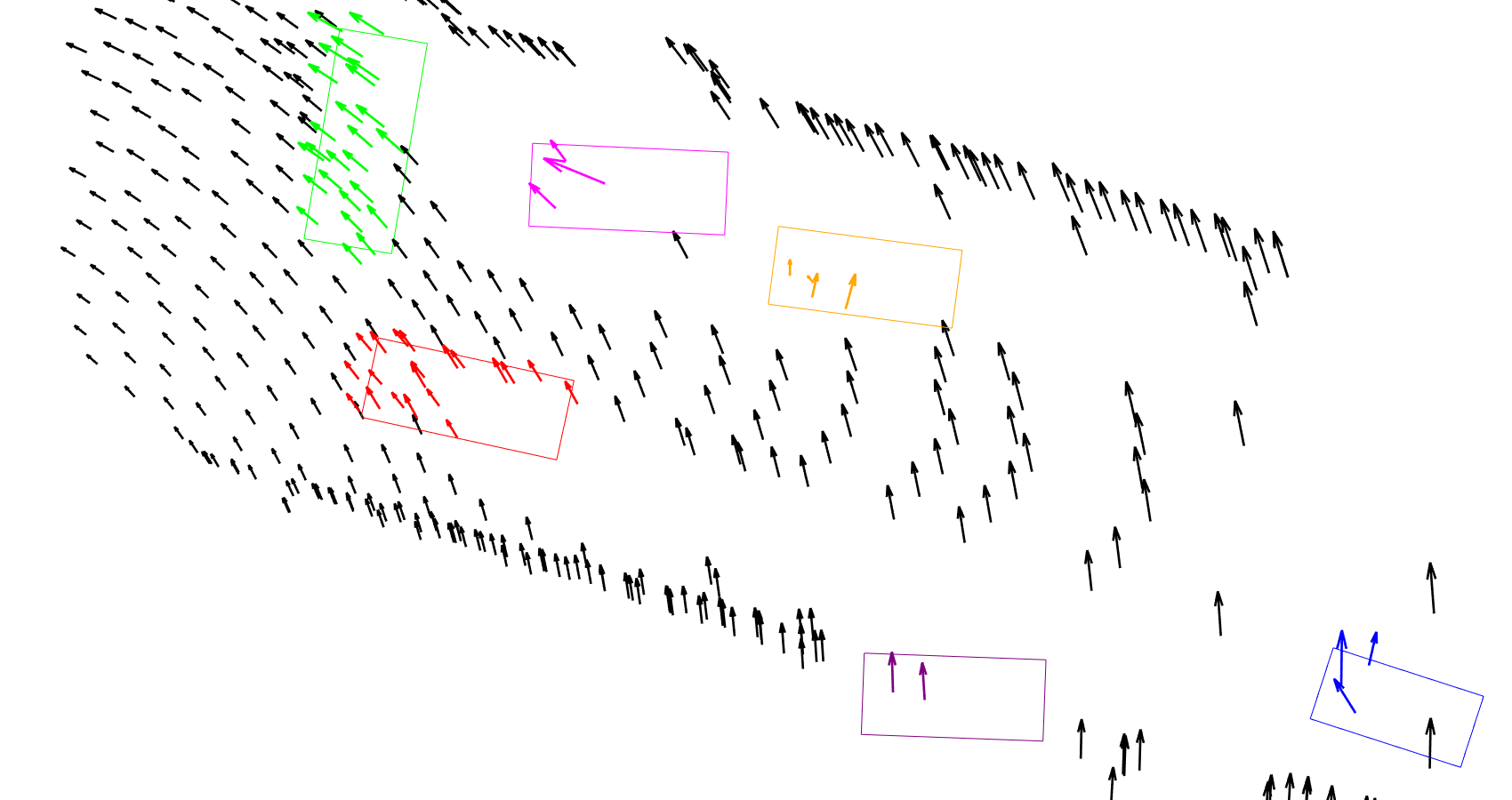}\\%
		(c) Dewan et al. ~\cite{Dewan2016IROS}+Det. & (d) ICP+Det.
	\end{tabular}
	\caption{
		{\bf Qualitative Comparison} of our method with the best performing baseline methods on an example from the Augmented KITTI. For clarity, we visualize only a subset of the points. Additional results can be found in the supplementary.
	}
	\label{fig:results}
	\vspace{-0.4cm}
\end{figure*}

\subsection{Baseline Methods}
We compare our method to four baselines: a point cloud-based method using a CRF~\cite{Dewan2016IROS}, two point-matching methods, and an Iterative Closest Point~\cite{Besl1992PAMI} (ICP) baseline.

\boldtext{Dewan et al.}\cite{Dewan2016IROS} estimate per-point rigid motion.
To arrive at object-level motion and ego-motion, we pool the estimates over our object detections and over the background.
As they only estimate valid scene flow for a subset of the points, we evaluate~\cite{Dewan2016IROS} only on those estimates and the comparison is therefore inherently biased in their favor.

\boldtext{Method Matching 3D Descriptors} yield a scene flow estimate for each point in the reference point cloud by finding correspondences of 3D features in two timesteps.
We evaluate two different descriptors: 3D Match~\cite{Zeng2017CVPR}, a learnable 3D descriptor trained on KITTI and Fast Point Feature Histogram features (FPFH)~\cite{Rusu2009ICRA}.
Based on the per-point scene flow, we fit rigid body motions to each of the objects and to the background, again using the object detections from our pipeline for a fair comparison.

\boldtext{Iterative Closest Point (ICP)}~\cite{Besl1992PAMI} outputs a transformation relating two point clouds to each other using an SVD-based point-to-point algorithm. 
We estimate object rigid motions by fitting the points of each detected 3D object in the first point cloud to the entire second point cloud.

\boldparagraph{Evaluation Metrics}
We quantify performance using several metrics applied to both the detected objects and the background.
To quantify the accuracy of the estimates independently from the detection accuracy, we only evaluate object motion on true positive detections.

\begin{itemize}
	\item For 3D scene flow, we use the average endpoint error between the prediction and the ground truth.
	\item Similarly, we list the average rotation and translation error averaged over all of the detected objects, and averaged over all scenes for the observer's ego-motion.
\end{itemize}

\subsection{Experimental Results}
\boldparagraph{The importance of simulated augmentation}
To quantify the value of our proposed LIDAR simulator for realistic augmentation with extra cars, we compare the performance of our method trained on the original KITTI object detection dataset with our method trained on both KITTI and Augmented KITTI.	
Table~\ref{tab:ablation} shows the results of this study. 
Our analysis shows that training using a combination of KITTI and augmented KITTI leads to significant performance gains, especially when evaluating on the more diverse vehicle motions in the validation set of Augmented KITTI.

\boldparagraph{Direct scene flow vs. object motion}
We have also evaluated the difference between estimating scene flow directly and calculating it from either dense or object-level rigid motion estimates. 
While scene flow computed from rigid motion estimates was qualitatively smoother, there was no significant difference in overall accuracy.

\boldparagraph{Comparison with the baselines}
Table~\ref{tab:results} summarizes the complete performance comparison on the KITTI test set.
Note that the comparison with Dewan et al.~\cite{Dewan2016IROS} is biased in their favor, as mentioned earlier, as we only evaluate their accuracy on the points they consider accurate.
Regardless, our method outperforms all baselines.
Additionally, we observe that the ICP-based method exhibits large errors for object motions.
This is because of objects with few points: ICP often performs very poorly on these, but while their impact on the dense evaluation is small they constitute a relatively larger fraction of the object-based evaluation. 
Visual examination (\figref{fig:results}) shows that the baseline methods predict a reasonable estimate for the background motion, but fail to estimate motion for dynamic objects; in contrast, our method is able to estimate these motions correctly.
This further reinforces the importance of training our method on scenes with many augmented cars and challenging and diverse motions.

Regarding execution time, our method requires 0.5 seconds to process one point cloud pair.
In comparison, Dewan et al. (4 seconds) and the 3D Match- and FPFH-based approaches (100 and 300 seconds, respectively) require significantly longer, while the ICP solution also takes 0.5 seconds but performs considerably worse.

\section{Conclusion}
\label{sec:conclusion}

In this paper, we have proposed a learning-based solution for estimating scene flow and rigid body motion from unstructured point clouds.
Our model simultaneously detects objects in the point clouds, estimates dense scene flow and rigid motion for all points in the cloud, and estimates object rigid motion for all detected objects as well as the observer.
We have shown that a global rigid motion representation is not amenable to fully convolutional estimation, and propose to use a local representation.
Our approach outperforms all evaluated baselines, yielding more accurate object motions in less time.


\section{Acknowledgements}
\label{sec:acknowledgements}
This work was supported by NVIDIA through its NVAIL program.

{\small
\bibliographystyle{ieee}
\bibliography{bibliography_long,bibliography,bibliography_custom}
}

\begin{appendices}
\section{Appendix}
\label{sec:appendix}
The following appendix provides additional qualitative comparisons of our method with the baselines methods along with examples of failure cases for our method. We also describe our workflow for generating the Augmented KITTI dataset here.
\section{Augmented KITTI}
\label{sec:aug_kitti}
\begin{figure*}[h]
	\centering
	\includegraphics[width=0.96\linewidth]{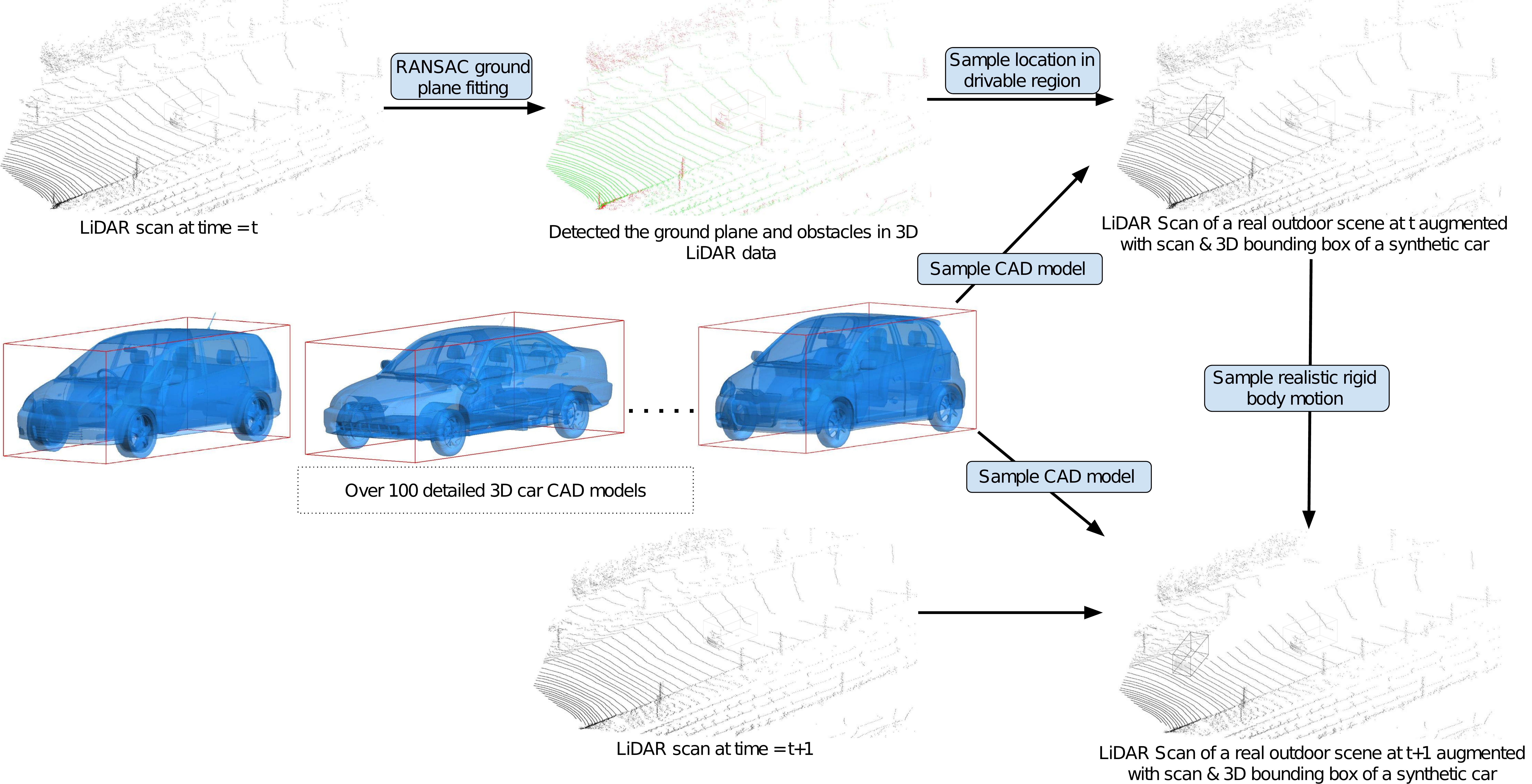}\\
	\caption{
		{\bf Augmented KITTI.} Workflow for generating the Augmented KITTI dataset.
	} 
	\label{fig:augKittiWorkflow}

\end{figure*}

Figure~\ref{fig:augKittiWorkflow} describes our workflow for generating the Augmented KITTI.
We start by fitting the ground plane using RANSAC 3D plane fitting; this allows us to detect obstacles and hence the drivable region.
In a second step, we randomly place virtual cars in the drivable region,  and simulate a new LIDAR scan that includes these virtual cars.
Our simulator uses a noise model learned from the real KITTI scanner, and also produces missing estimates at transparent surfaces using the transparency information provided by the CAD models.
Additionally, we remove points in the original scan which become occluded by the augmented car by tracing a ray between each point and the LIDAR, and removing those points whose ray intersects with the car mesh.
Finally, we sample the augmented car's rigid motion using a simple approximation of the Ackermann steering geometry, place the car at the corresponding location in the next frame, and repeat the LIDAR simulation.

\section{Qualitative Comparison to Baseline Methods}
\label{sec:add_results}

\begin{figure*}[ht!]
	\setlength\tabcolsep{1pt}
	\def\arraystretch{1}
	\def\imgw{\textwidth}
	\begin{tabular}{c}
		\includegraphics[width=\imgw]{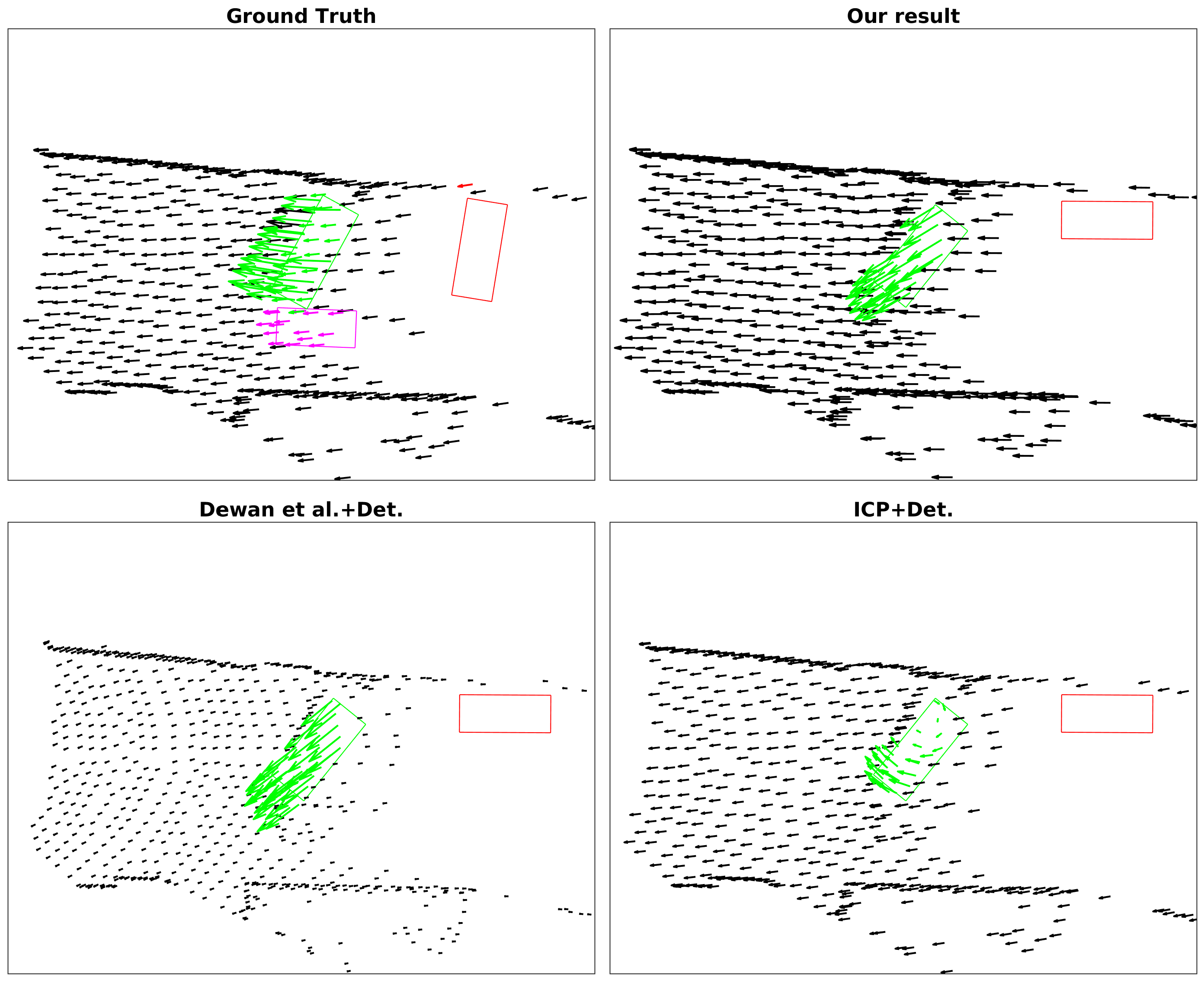}
	\end{tabular}
	\caption{
		{\bf Qualitative Comparison} of our method with the best performing baseline methods on an example from the test set of the Augmented KITTI dataset. We observe here that our method predicts wrong scene flow for points on the green car in the reference point cloud at frame $t$ by matching them with points on the other car (pink) in close proximity at frame $t+1$. For clarity, we visualize only a subset of the points.}
	\label{fig:results20}
	\vspace{-0.4cm}
\end{figure*}

\begin{figure*}[ht!]
	\setlength\tabcolsep{1pt}
	\def\arraystretch{1}
	\def\imgw{\textwidth}
	\begin{tabular}{c}
		\includegraphics[width=\imgw]{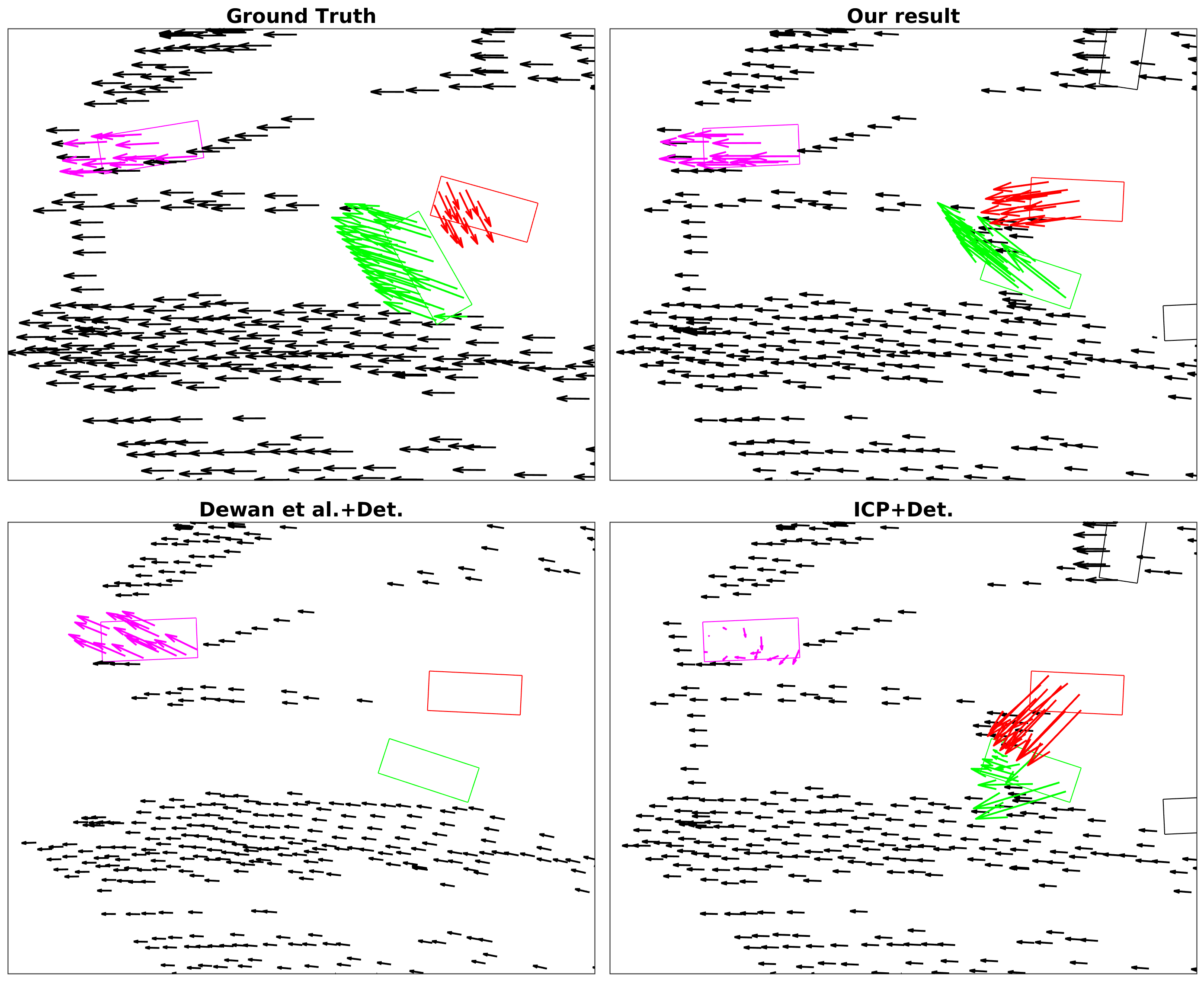}
	\end{tabular}
	\caption{
		{\bf Qualitative Comparison} of our method with the best performing baseline methods on an example from the test set of the Augmented KITTI dataset. We observe here that our method predicts wrong scene flow for points on the red car in the reference point cloud at frame $t$ by matching them with points on the other car (green) in close proximity at frame $t+1$. For clarity, we visualize only a subset of the points.}
	\label{fig:results21}
	\vspace{-0.4cm}
\end{figure*}

\begin{figure*}[ht!]
	\setlength\tabcolsep{1pt}
	\def\arraystretch{1}
	\def\imgw{\textwidth}
	\begin{tabular}{c}
		\includegraphics[width=\imgw]{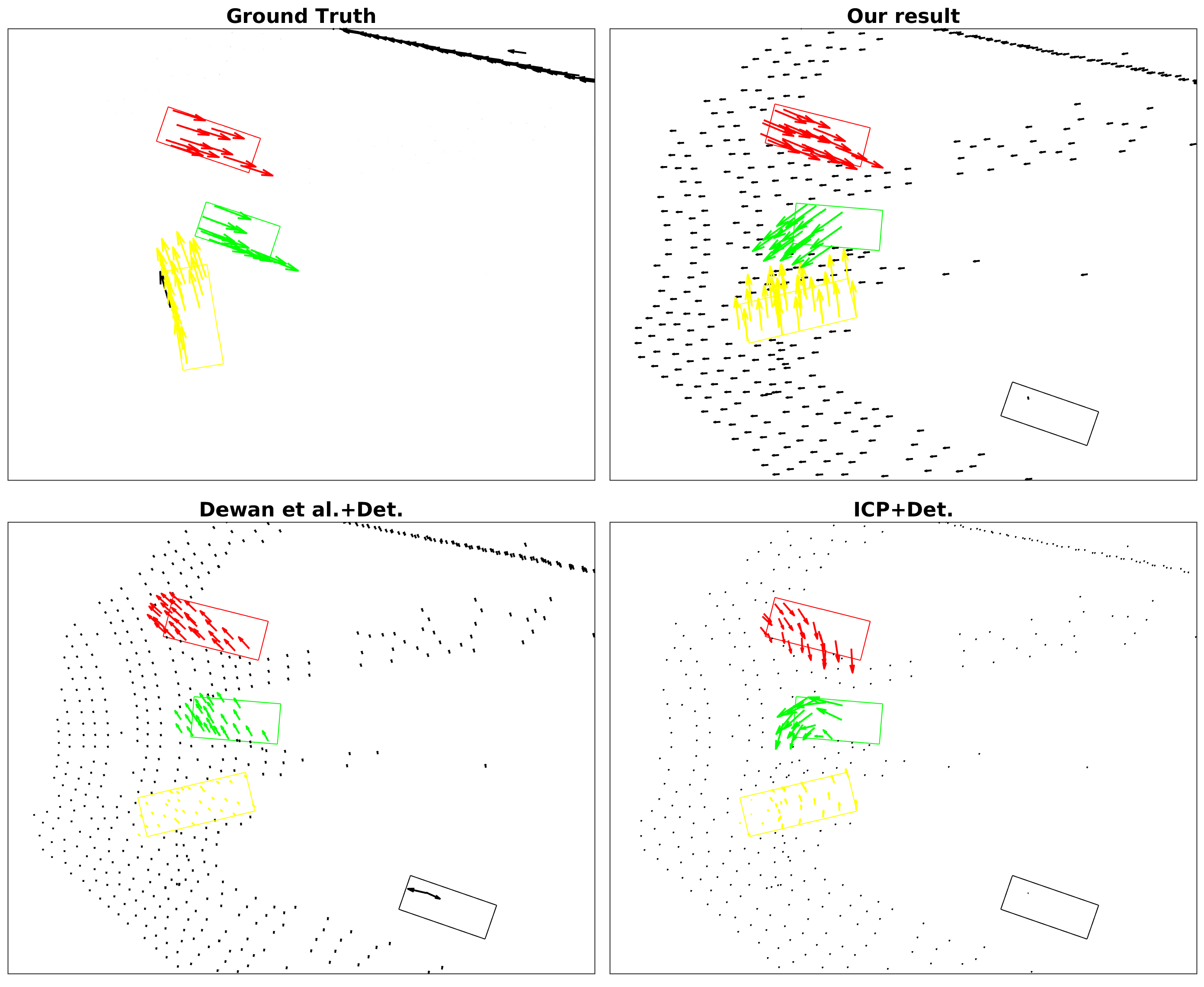}
	\end{tabular}
	\caption{
		{\bf Qualitative Comparison} of our method with the best performing baseline methods on an example from the test set of the Augmented KITTI dataset. We observe here that our method predicts wrong scene flow for points on the green car in the reference point cloud at frame $t$ by matching them with points on the other car (yellow) in close proximity at frame $t+1$. For clarity, we visualize only a subset of the points.}
	\label{fig:results22}
	\vspace{-0.4cm}
\end{figure*}

\newcommand{\capqual}{{\bf Qualitative Comparison} of our method with the best performing baseline methods on an example from the test set of the Augmented KITTI dataset. For clarity, we visualize only a subset of the points.}
Figures~\ref{fig:results1} - \ref{fig:results22} show qualitative comparison of our method with the best performing baseline methods on examples from the test set of the Augmented KITTI dataset. The qualitative results show that our method predicts motion for both background and foreground parts of the scene with higher accuracy than all the baselines on a diverse range of scenes and motions. 

In Figures~\ref{fig:results20} - \ref{fig:results22}, we provide challenging examples where our method fails to predict the correct scene flow. We observe here that in case of scenes with two or more cars in very close proximity, our method may predict wrong scene flow for points on one car in the reference point cloud at frame $t$ by matching them with points on the other car in close proximity at frame $t+1$. However, we note  that, even for these failure cases our method performs better than the baseline methods.

\begin{figure*}[ht!]
	\setlength\tabcolsep{1pt}
	\def\arraystretch{1}
	\def\imgw{\textwidth}
	\begin{tabular}{c}
		\includegraphics[width=\imgw]{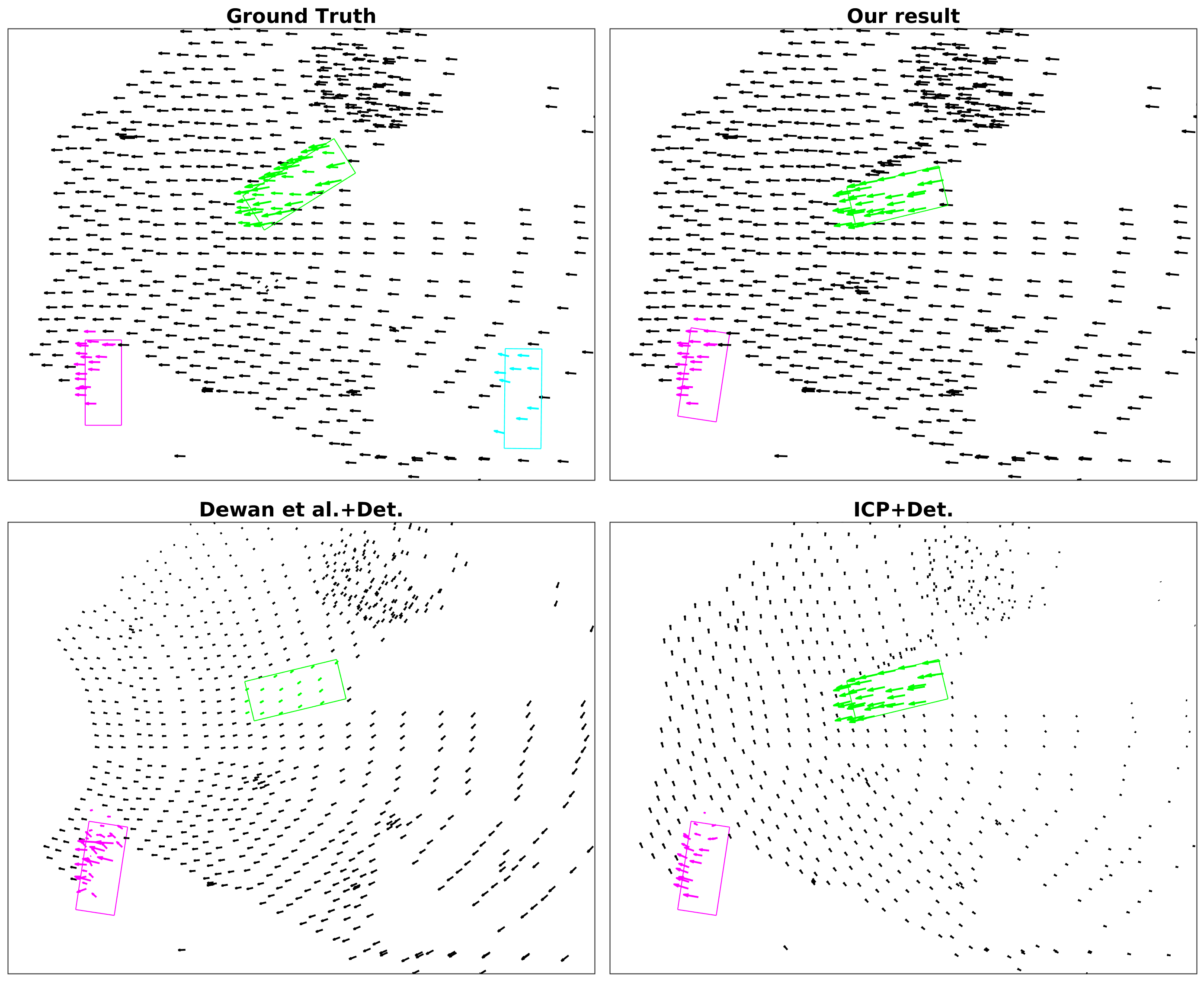}
	\end{tabular}
	\caption{
		{\capqual}}
	\label{fig:results1}
	\vspace{-0.4cm}
\end{figure*}
\begin{figure*}[ht!]
	\setlength\tabcolsep{1pt}
	\def\arraystretch{1}
	\def\imgw{\textwidth}
	\begin{tabular}{c}
		\includegraphics[width=\imgw]{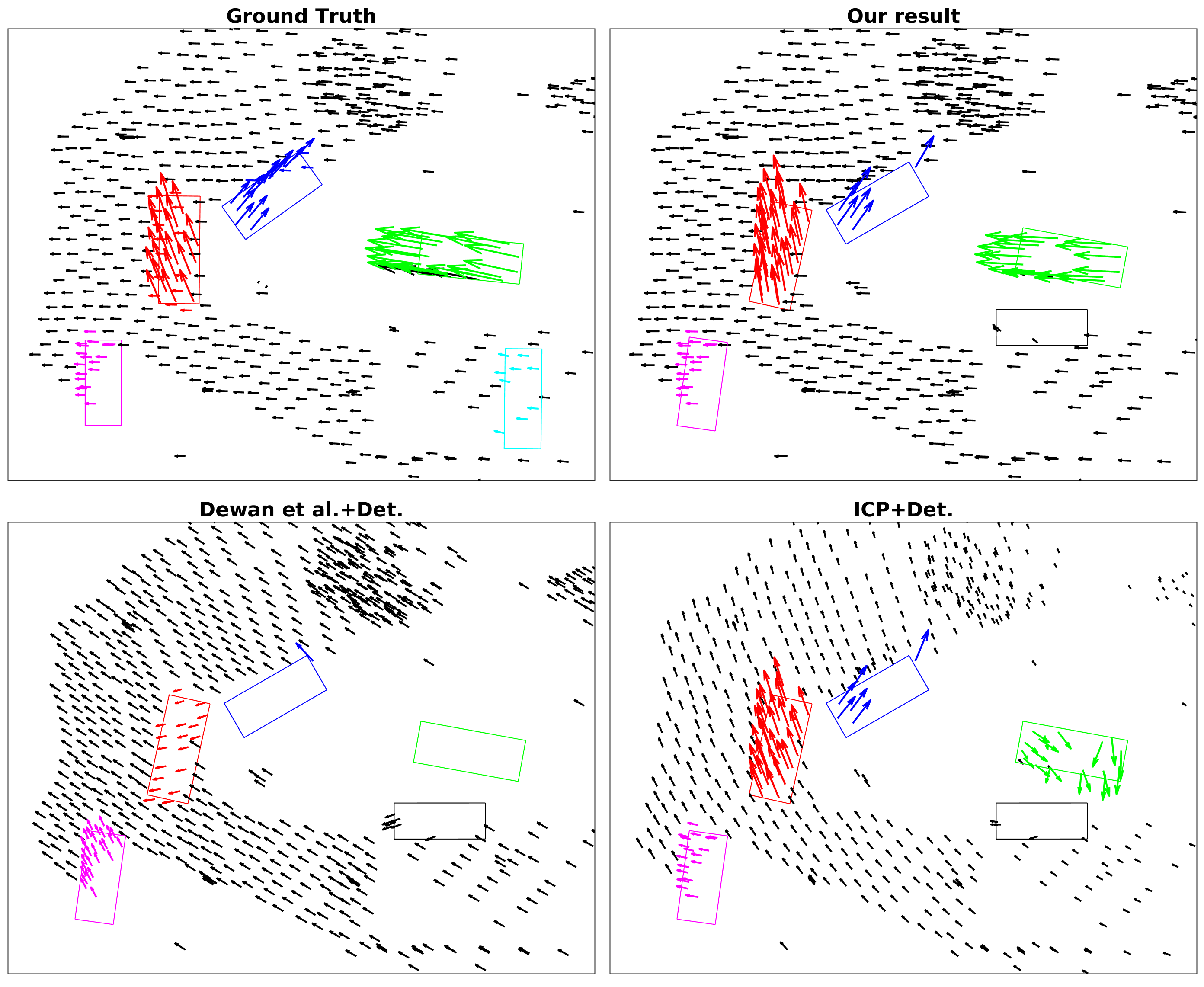}
	\end{tabular}
	\caption{
		{\capqual}}
	\label{fig:results2}
	\vspace{-0.4cm}
\end{figure*}
\begin{figure*}[ht!]
	\setlength\tabcolsep{1pt}
	\def\arraystretch{1}
	\def\imgw{\textwidth}
	\begin{tabular}{c}
		\includegraphics[width=\imgw]{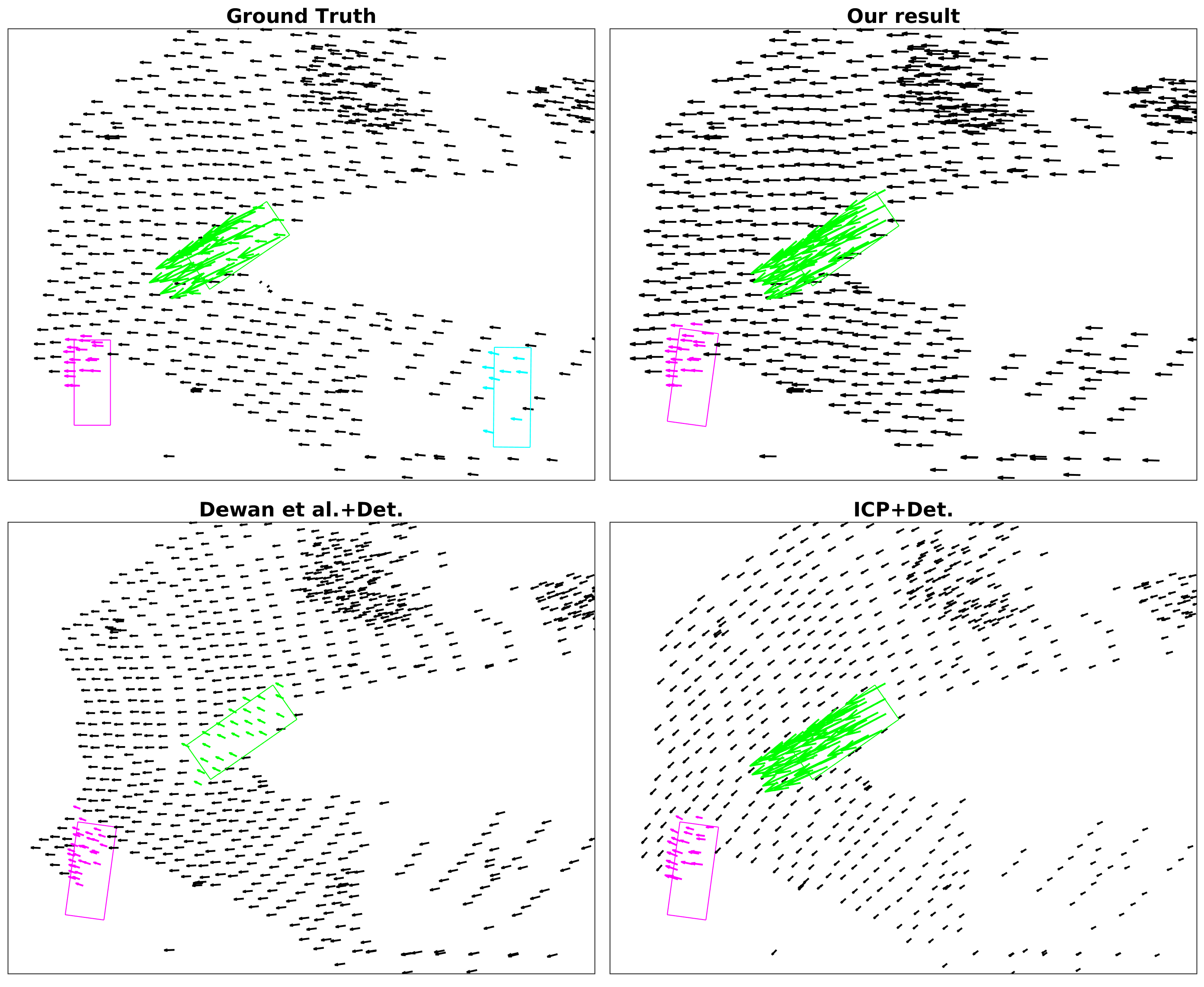}
	\end{tabular}
	\caption{
		{\capqual}}
	\label{fig:results3}
	\vspace{-0.4cm}
\end{figure*}
\begin{figure*}[ht!]
	\setlength\tabcolsep{1pt}
	\def\arraystretch{1}
	\def\imgw{\textwidth}
	\begin{tabular}{c}
		\includegraphics[width=\imgw]{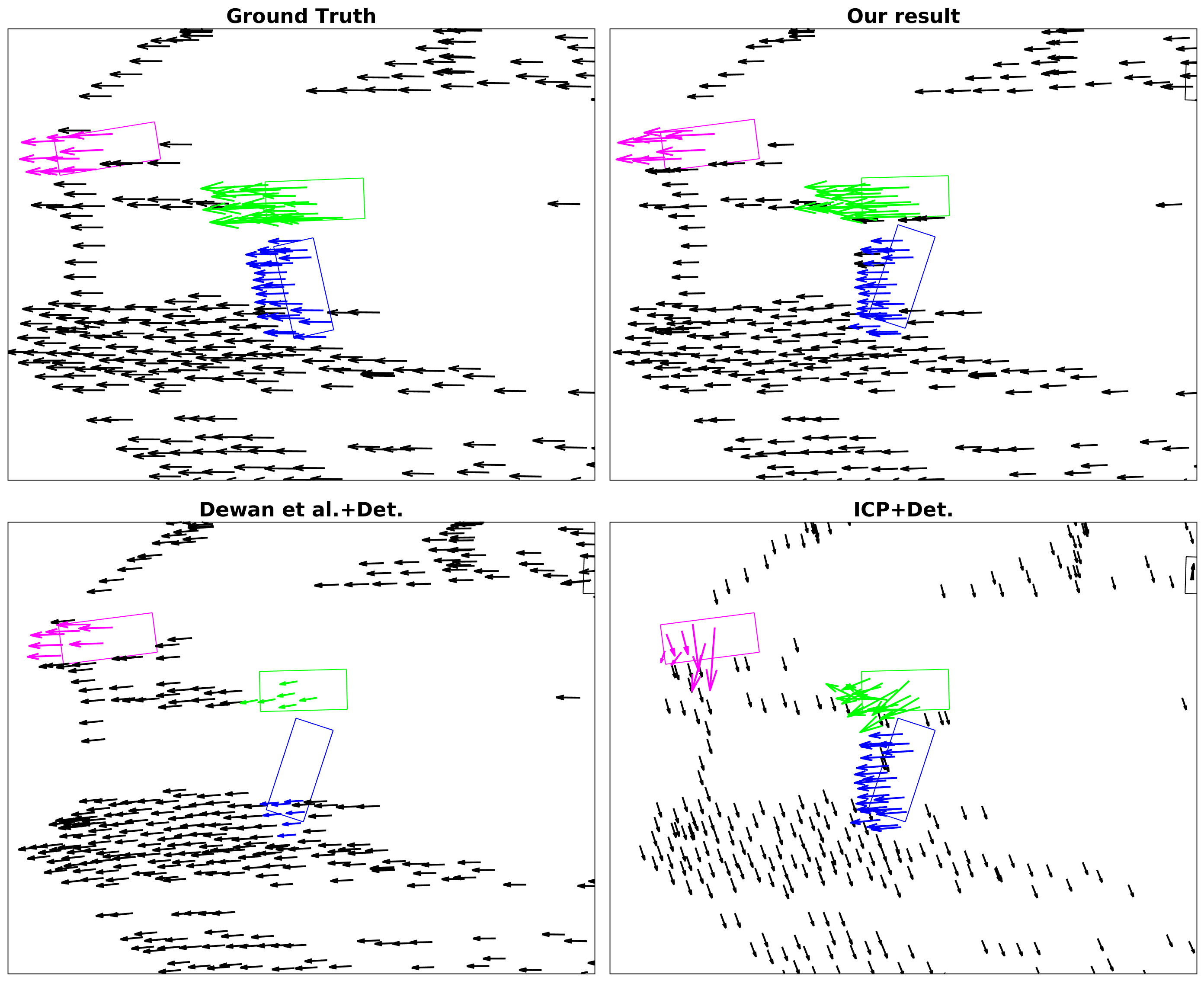}
	\end{tabular}
	\caption{
		{\capqual}}
	\label{fig:results4}
	\vspace{-0.4cm}
\end{figure*}
\begin{figure*}[ht!]
	\setlength\tabcolsep{1pt}
	\def\arraystretch{1}
	\def\imgw{\textwidth}
	\begin{tabular}{c}
		\includegraphics[width=\imgw]{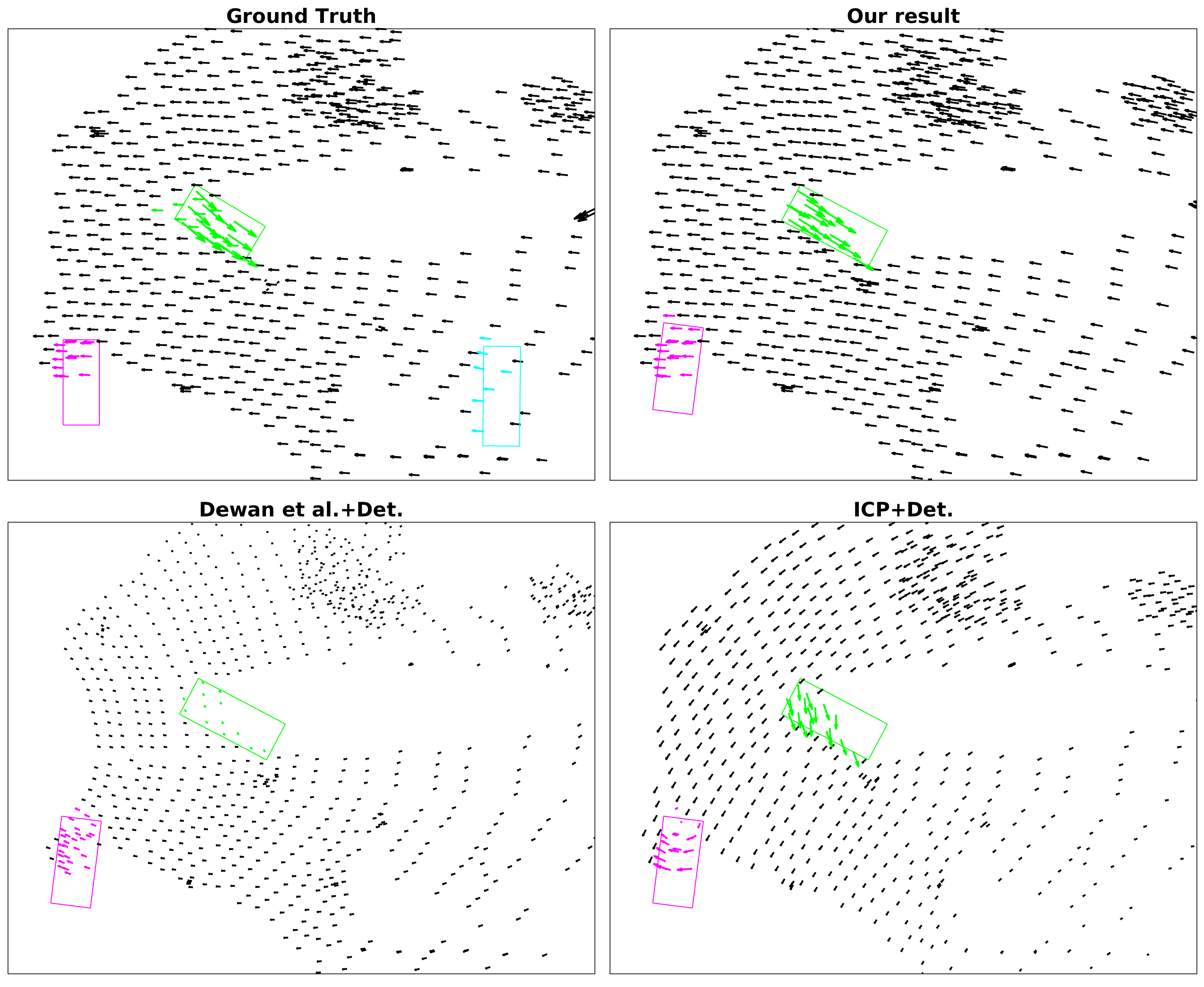}
	\end{tabular}
	\caption{
		{\capqual}}
	\label{fig:results5}
	\vspace{-0.4cm}
\end{figure*}
\begin{figure*}[ht!]
	\setlength\tabcolsep{1pt}
	\def\arraystretch{1}
	\def\imgw{\textwidth}
	\begin{tabular}{c}
		\includegraphics[width=\imgw]{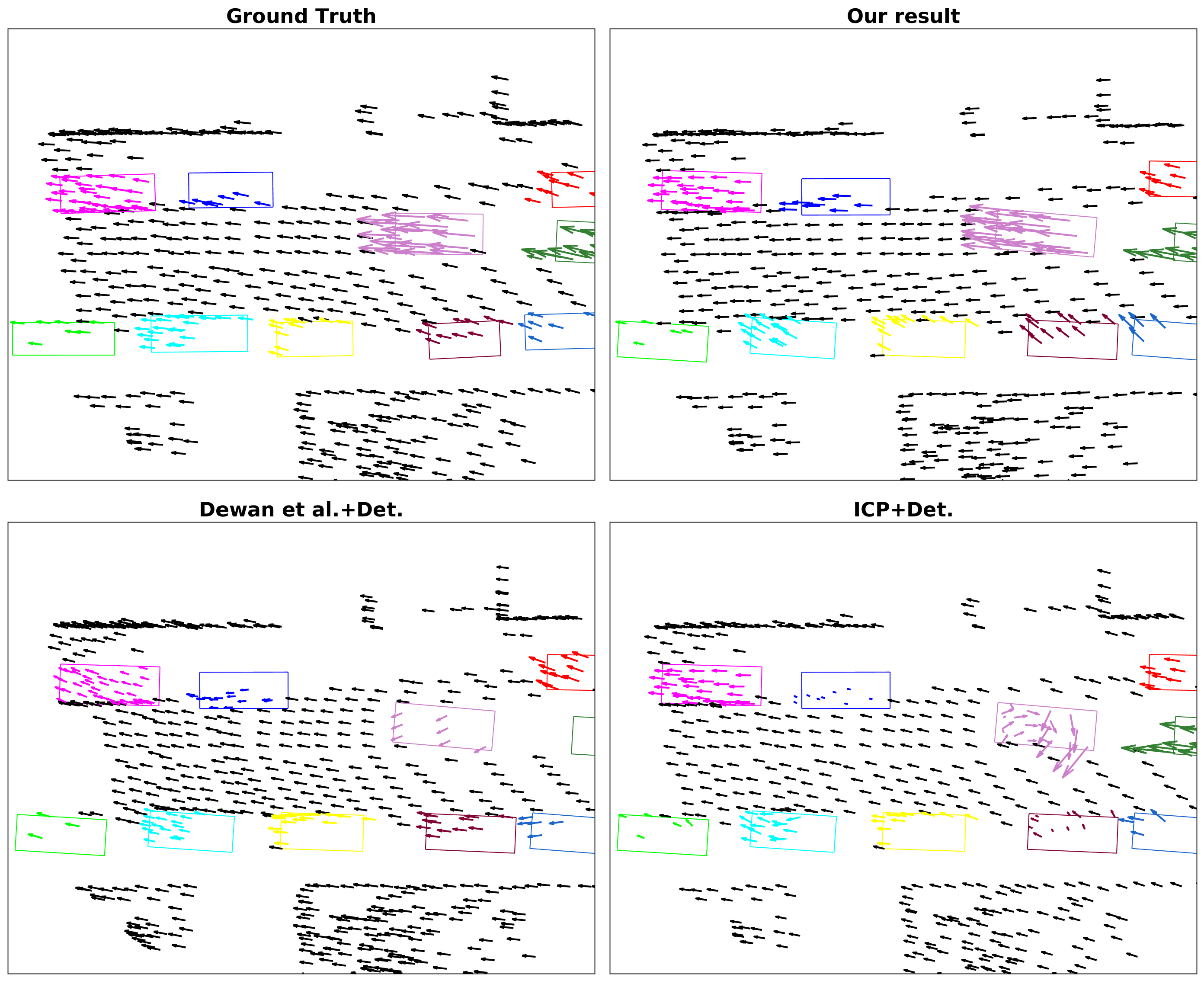}
	\end{tabular}
	\caption{
		{\capqual}}
	\label{fig:results6}
	\vspace{-0.4cm}
\end{figure*}
\begin{figure*}[ht!]
	\setlength\tabcolsep{1pt}
	\def\arraystretch{1}
	\def\imgw{\textwidth}
	\begin{tabular}{c}
		\includegraphics[width=\imgw]{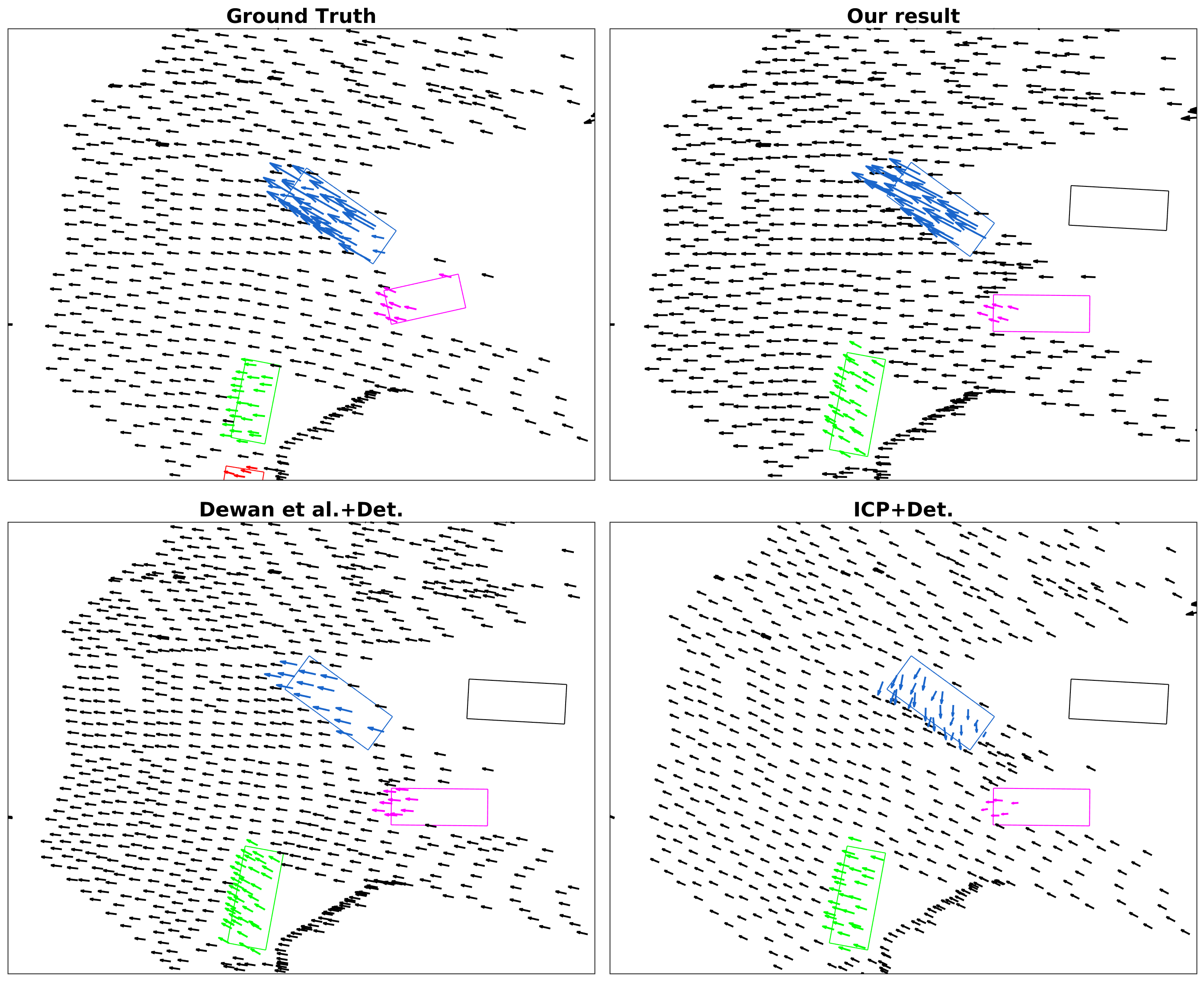}
	\end{tabular}
	\caption{
		{\capqual}}
	\label{fig:results7}
	\vspace{-0.4cm}
\end{figure*}
\begin{figure*}[ht!]
	\setlength\tabcolsep{1pt}
	\def\arraystretch{1}
	\def\imgw{\textwidth}
	\begin{tabular}{c}
		\includegraphics[width=\imgw]{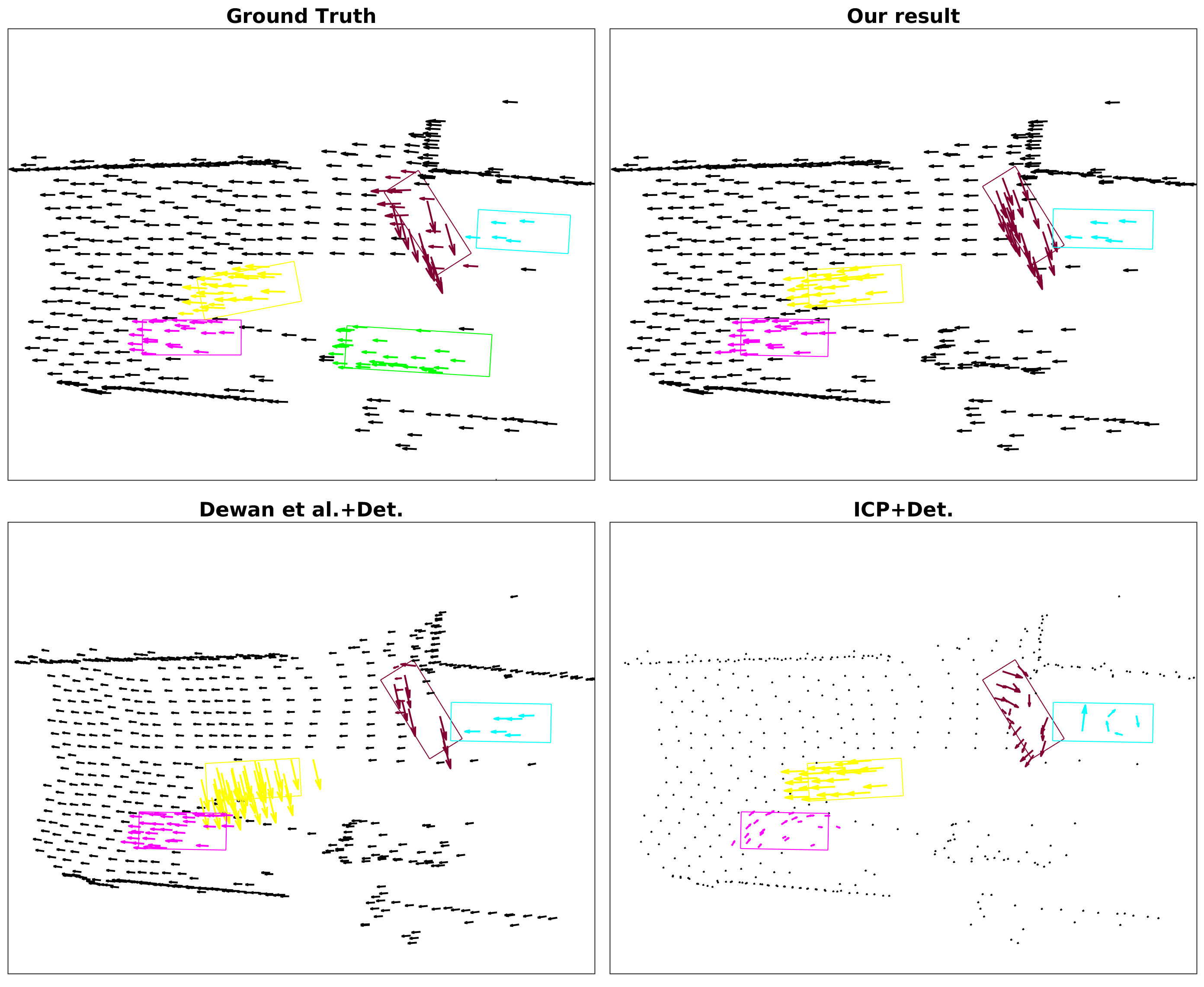}
	\end{tabular}
	\caption{
		{\capqual}}
	\label{fig:results8}
	\vspace{-0.4cm}
\end{figure*}
\begin{figure*}[ht!]
	\setlength\tabcolsep{1pt}
	\def\arraystretch{1}
	\def\imgw{\textwidth}
	\begin{tabular}{c}
		\includegraphics[width=\imgw]{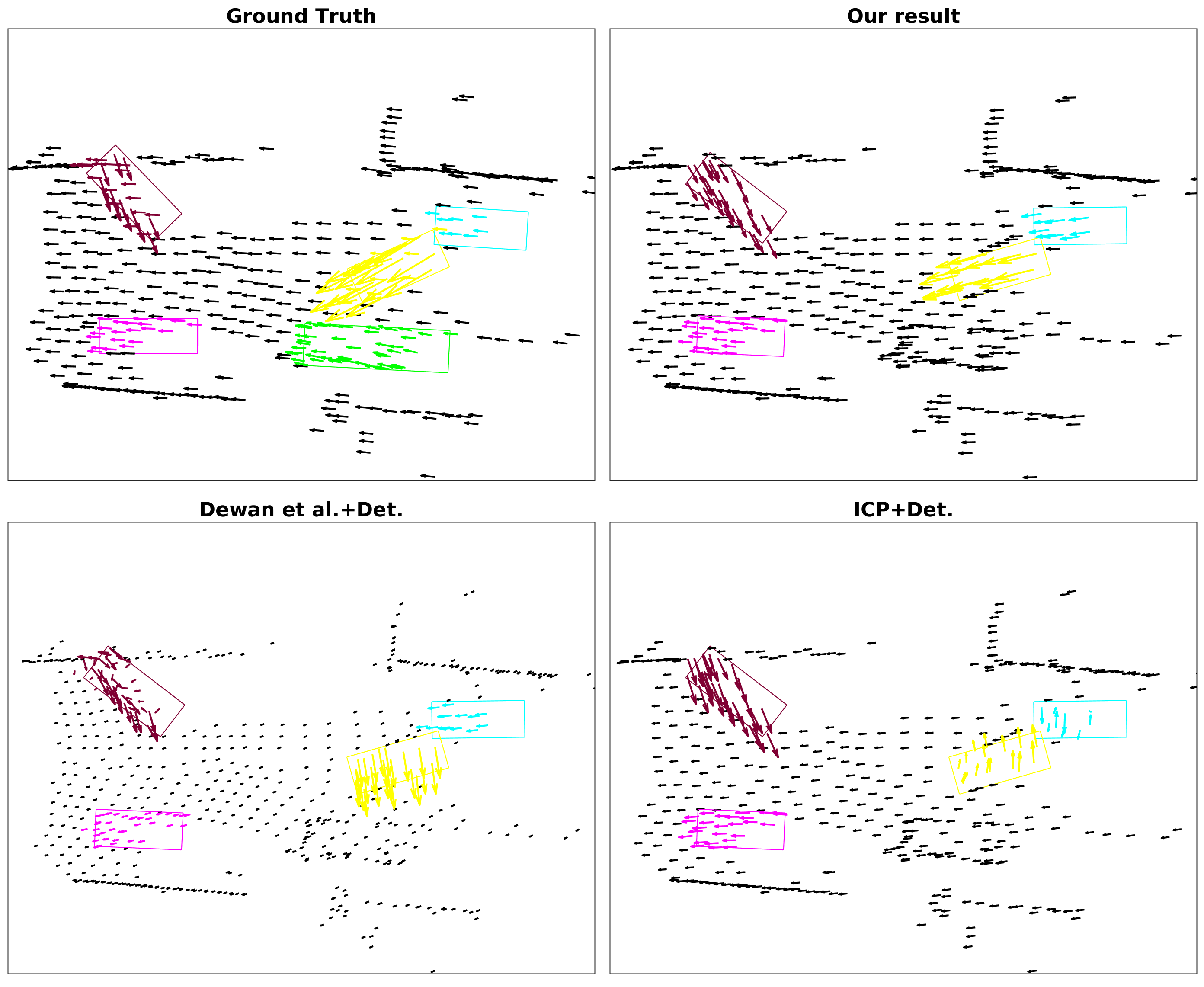}
	\end{tabular}
	\caption{
		{\capqual}}
	\label{fig:results9}
	\vspace{-0.4cm}
\end{figure*}
\begin{figure*}[ht!]
	\setlength\tabcolsep{1pt}
	\def\arraystretch{1}
	\def\imgw{\textwidth}
	\begin{tabular}{c}
		\includegraphics[width=\imgw]{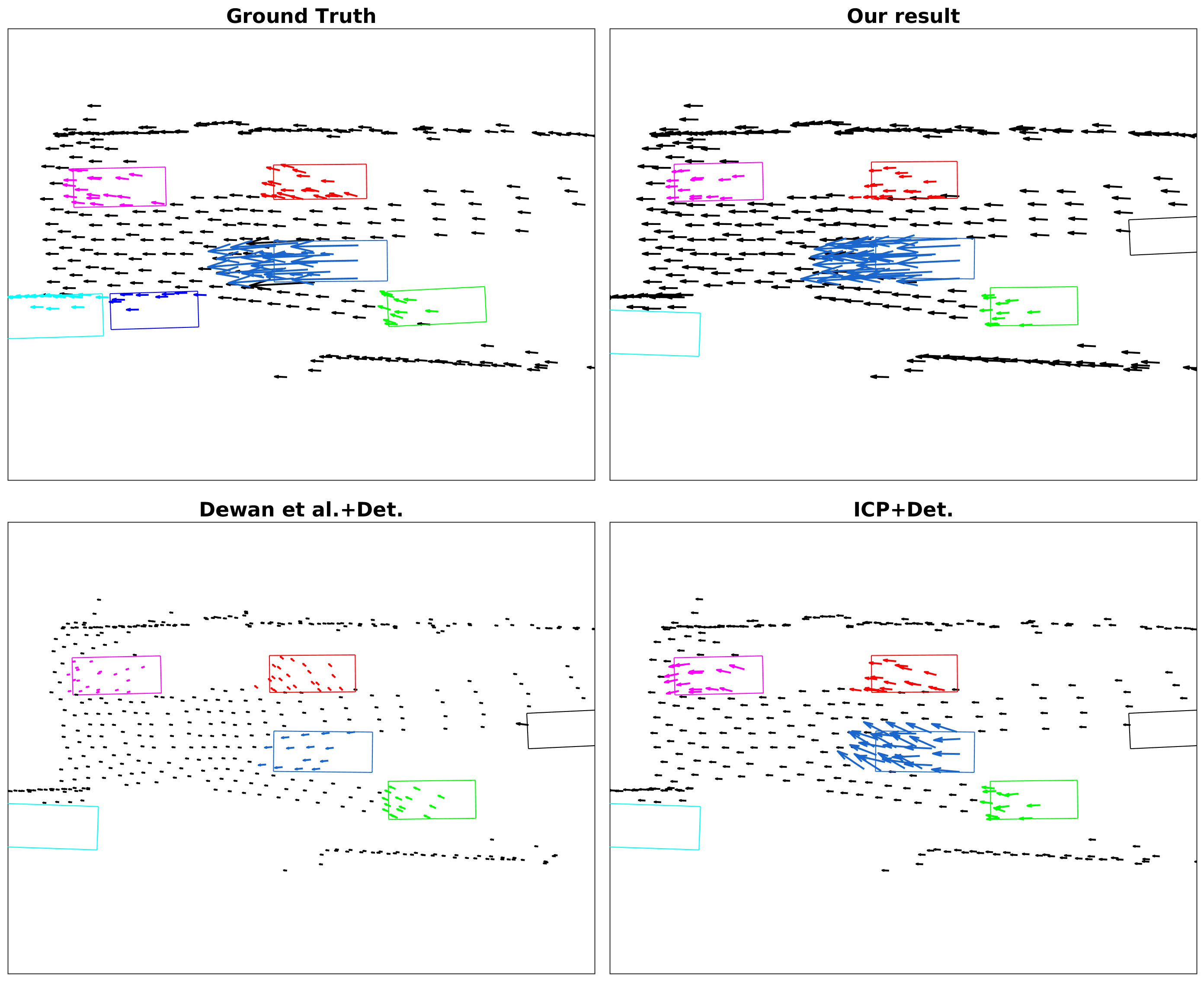}
	\end{tabular}
	\caption{
		{\capqual}}
	\label{fig:results10}
	\vspace{-0.4cm}
\end{figure*}
\begin{figure*}[ht!]
	\setlength\tabcolsep{1pt}
	\def\arraystretch{1}
	\def\imgw{\textwidth}
	\begin{tabular}{c}
		\includegraphics[width=\imgw]{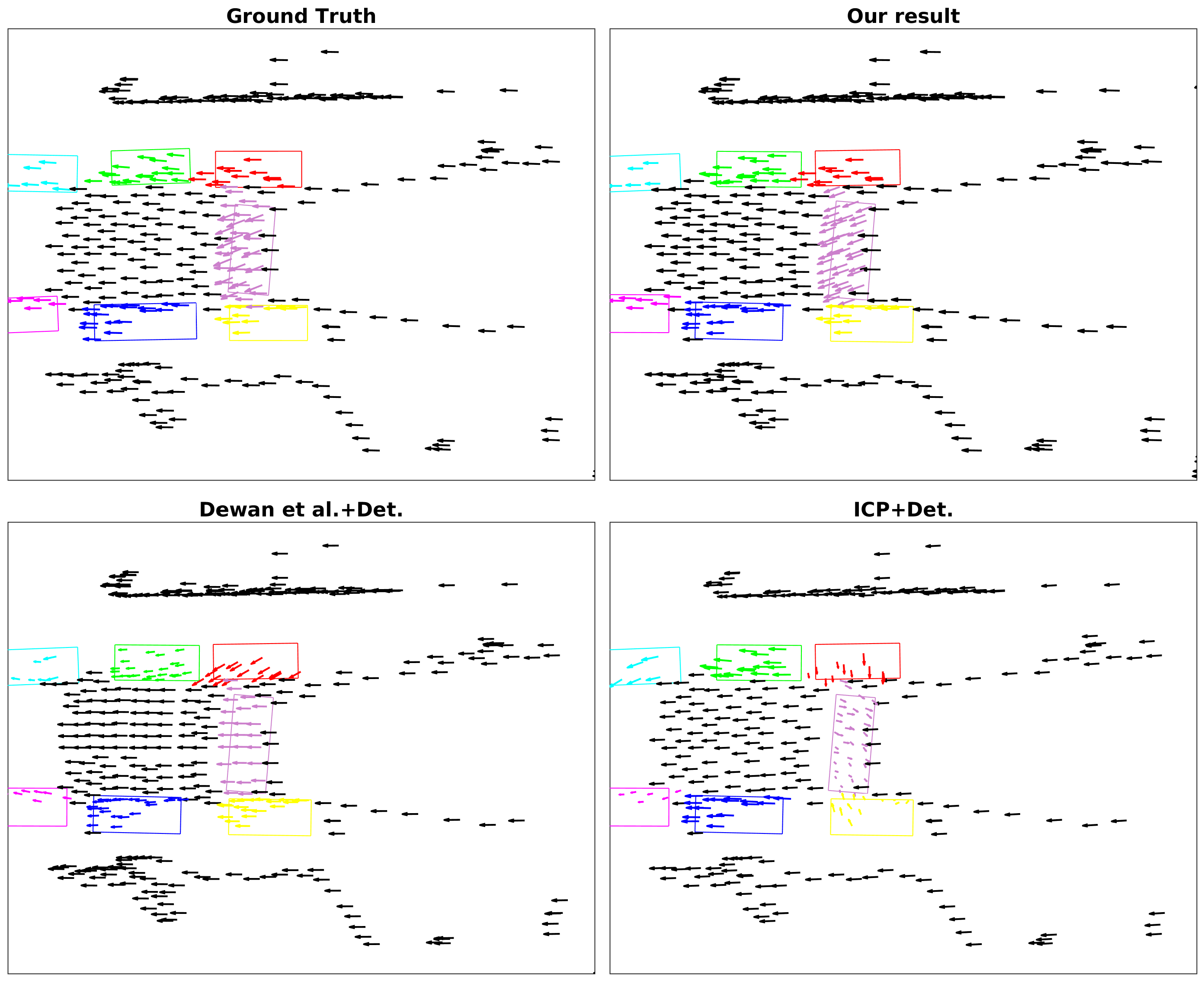}
	\end{tabular}
	\caption{
		{\capqual}}
	\label{fig:results11}
	\vspace{-0.4cm}
\end{figure*}
\begin{figure*}[ht!]
	\setlength\tabcolsep{1pt}
	\def\arraystretch{1}
	\def\imgw{\textwidth}
	\begin{tabular}{c}
		\includegraphics[width=\imgw]{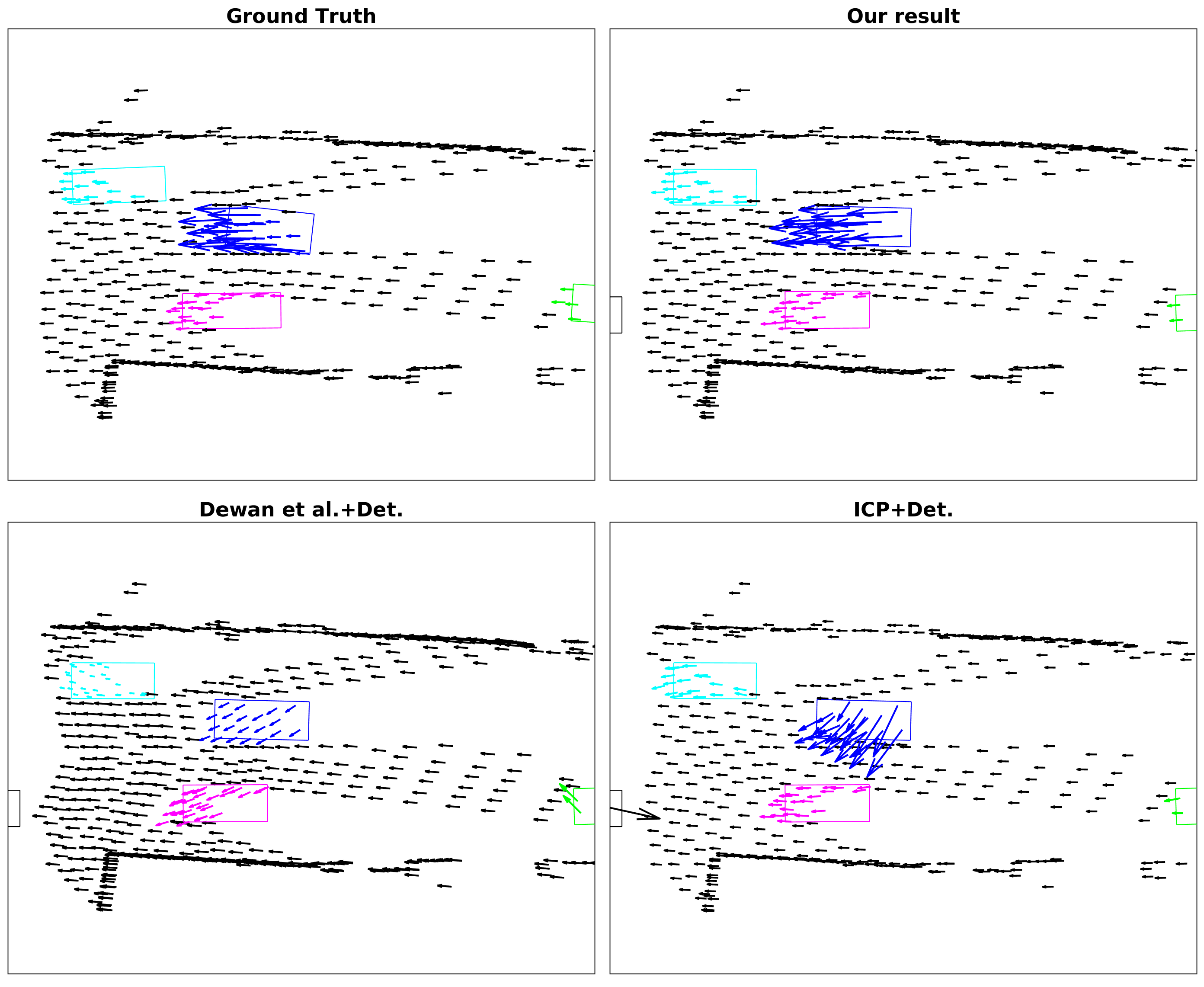}
	\end{tabular}
	\caption{
		{\capqual}}
	\label{fig:results12}
	\vspace{-0.4cm}
\end{figure*}
\begin{figure*}[ht!]
	\setlength\tabcolsep{1pt}
	\def\arraystretch{1}
	\def\imgw{\textwidth}
	\begin{tabular}{c}
		\includegraphics[width=\imgw]{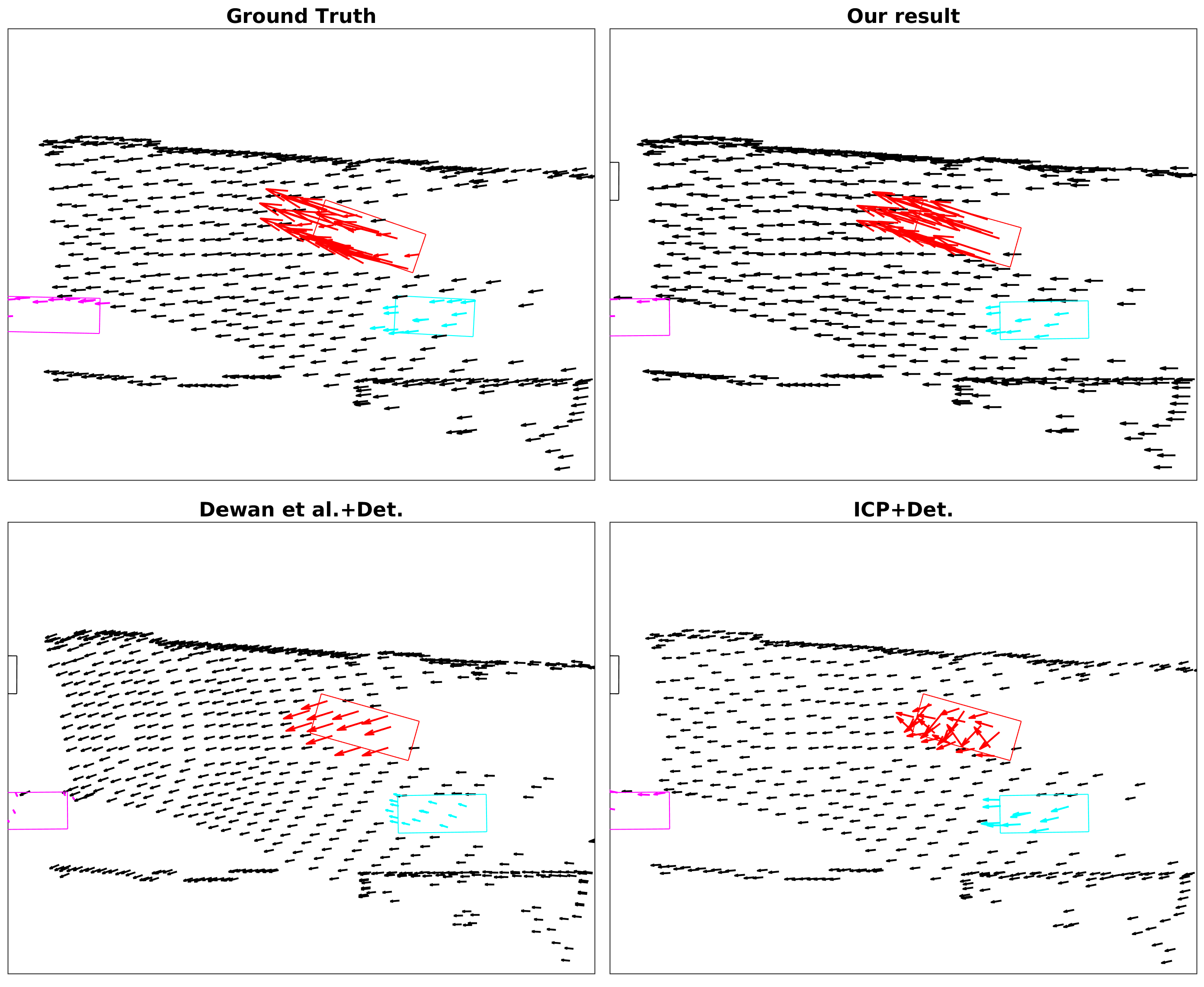}
	\end{tabular}
	\caption{
		{\capqual}}
	\label{fig:results13}
	\vspace{-0.4cm}
\end{figure*}
\begin{figure*}[ht!]
	\setlength\tabcolsep{1pt}
	\def\arraystretch{1}
	\def\imgw{\textwidth}
	\begin{tabular}{c}
		\includegraphics[width=\imgw]{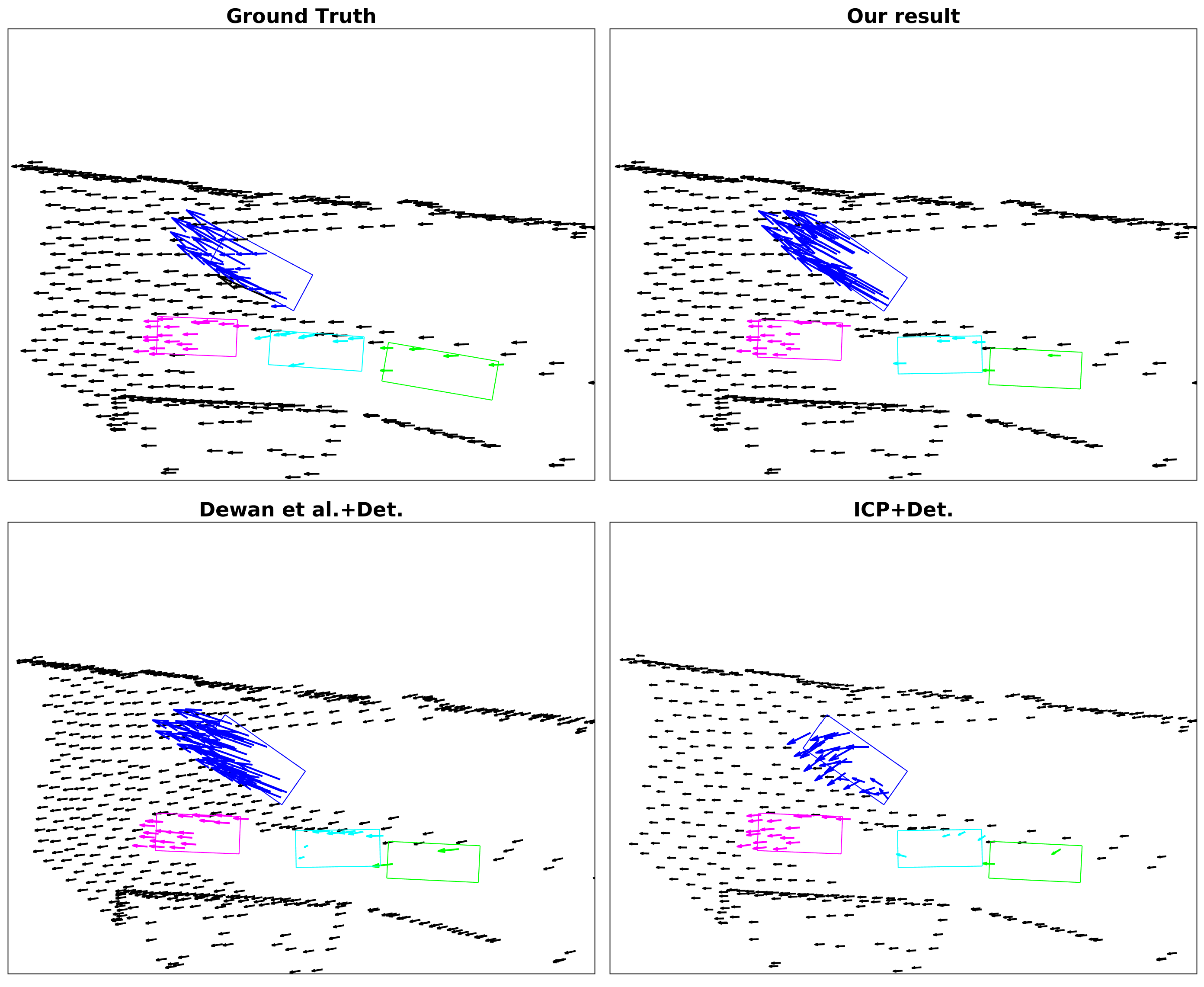}
	\end{tabular}
	\caption{
		{\capqual}}
	\label{fig:results14}
	\vspace{-0.4cm}
\end{figure*}
\begin{figure*}[ht!]
	\setlength\tabcolsep{1pt}
	\def\arraystretch{1}
	\def\imgw{\textwidth}
	\begin{tabular}{c}
		\includegraphics[width=\imgw]{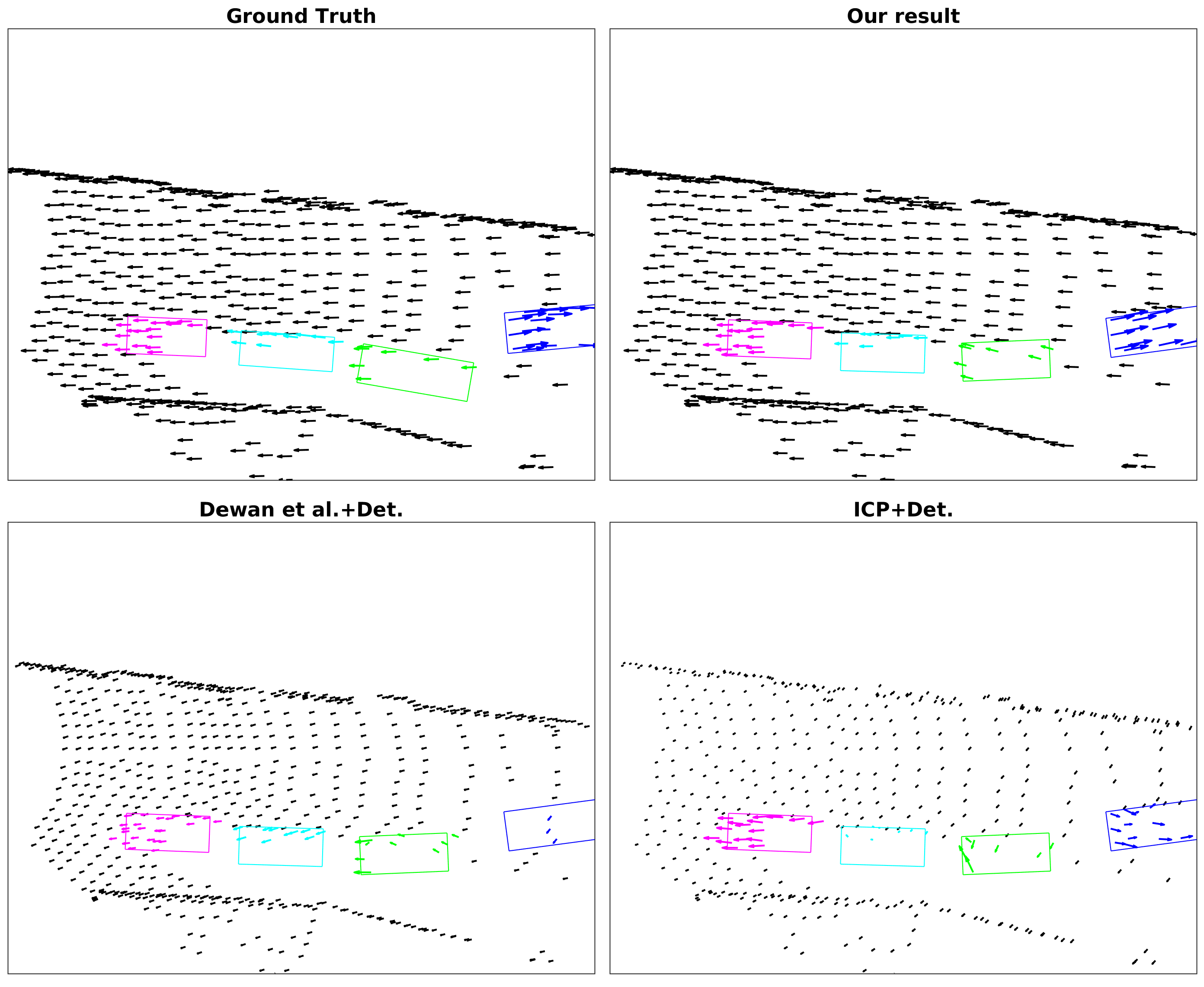}
	\end{tabular}
	\caption{
		{\capqual}}
	\label{fig:results15}
	\vspace{-0.4cm}
\end{figure*}
\begin{figure*}[ht!]
	\setlength\tabcolsep{1pt}
	\def\arraystretch{1}
	\def\imgw{\textwidth}
	\begin{tabular}{c}
		\includegraphics[width=\imgw]{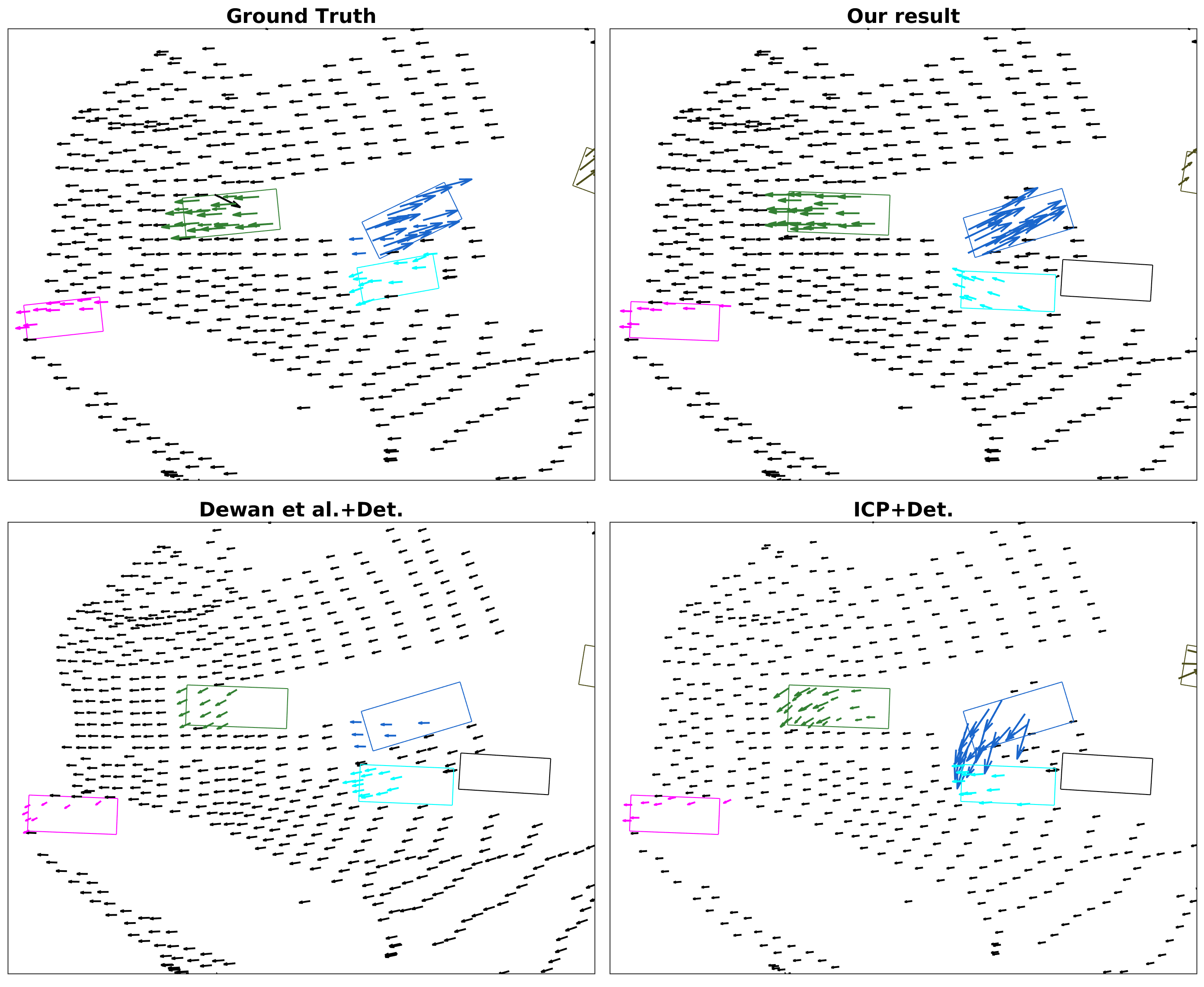}
	\end{tabular}
	\caption{
		{\capqual}}
	\label{fig:results16}
	\vspace{-0.4cm}
\end{figure*}
\begin{figure*}[ht!]
	\setlength\tabcolsep{1pt}
	\def\arraystretch{1}
	\def\imgw{\textwidth}
	\begin{tabular}{c}
		\includegraphics[width=\imgw]{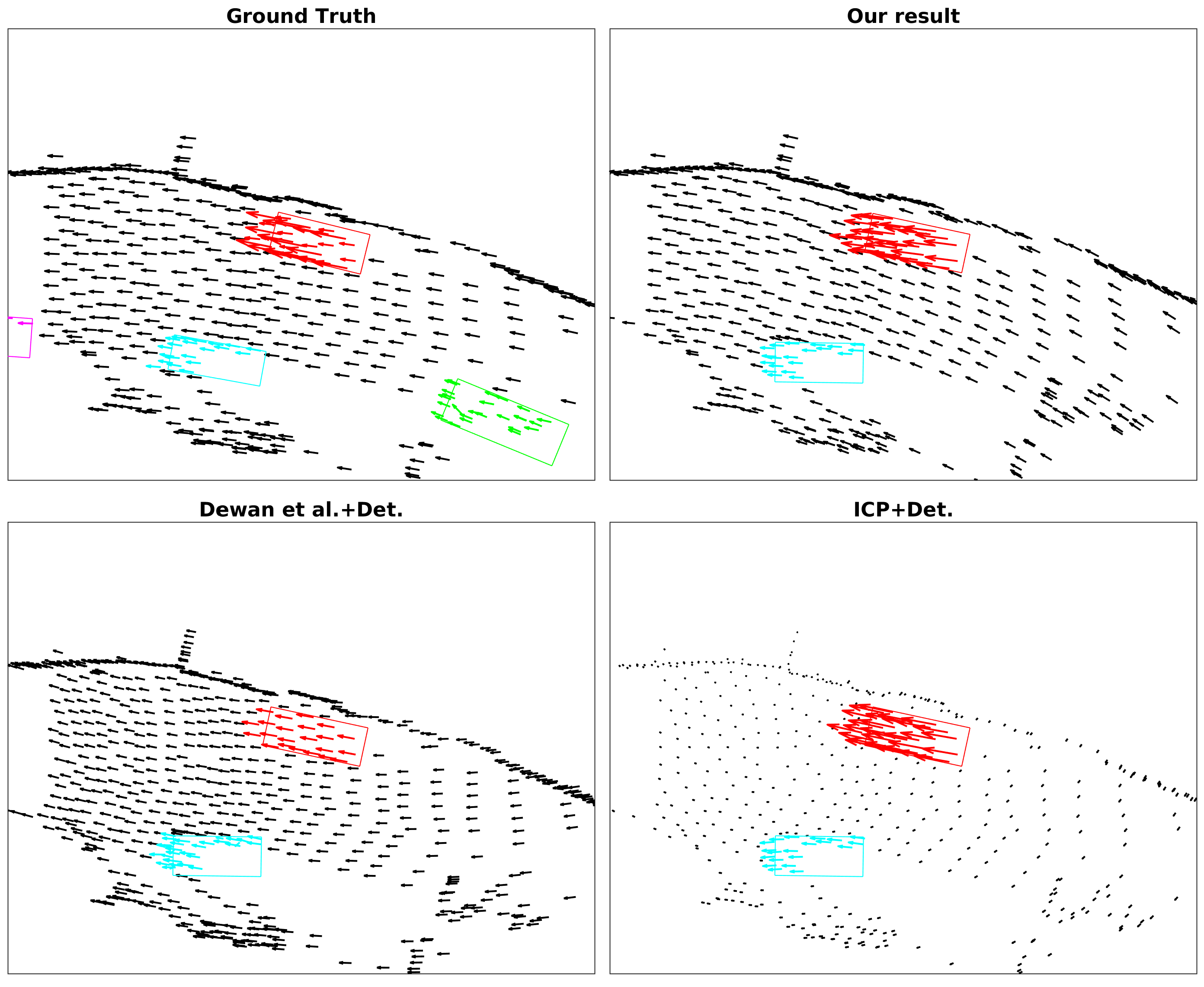}
	\end{tabular}
	\caption{
		{\capqual}}
	\label{fig:results17}
	\vspace{-0.4cm}
\end{figure*}
\begin{figure*}[ht!]
	\setlength\tabcolsep{1pt}
	\def\arraystretch{1}
	\def\imgw{\textwidth}
	\begin{tabular}{c}
		\includegraphics[width=\imgw]{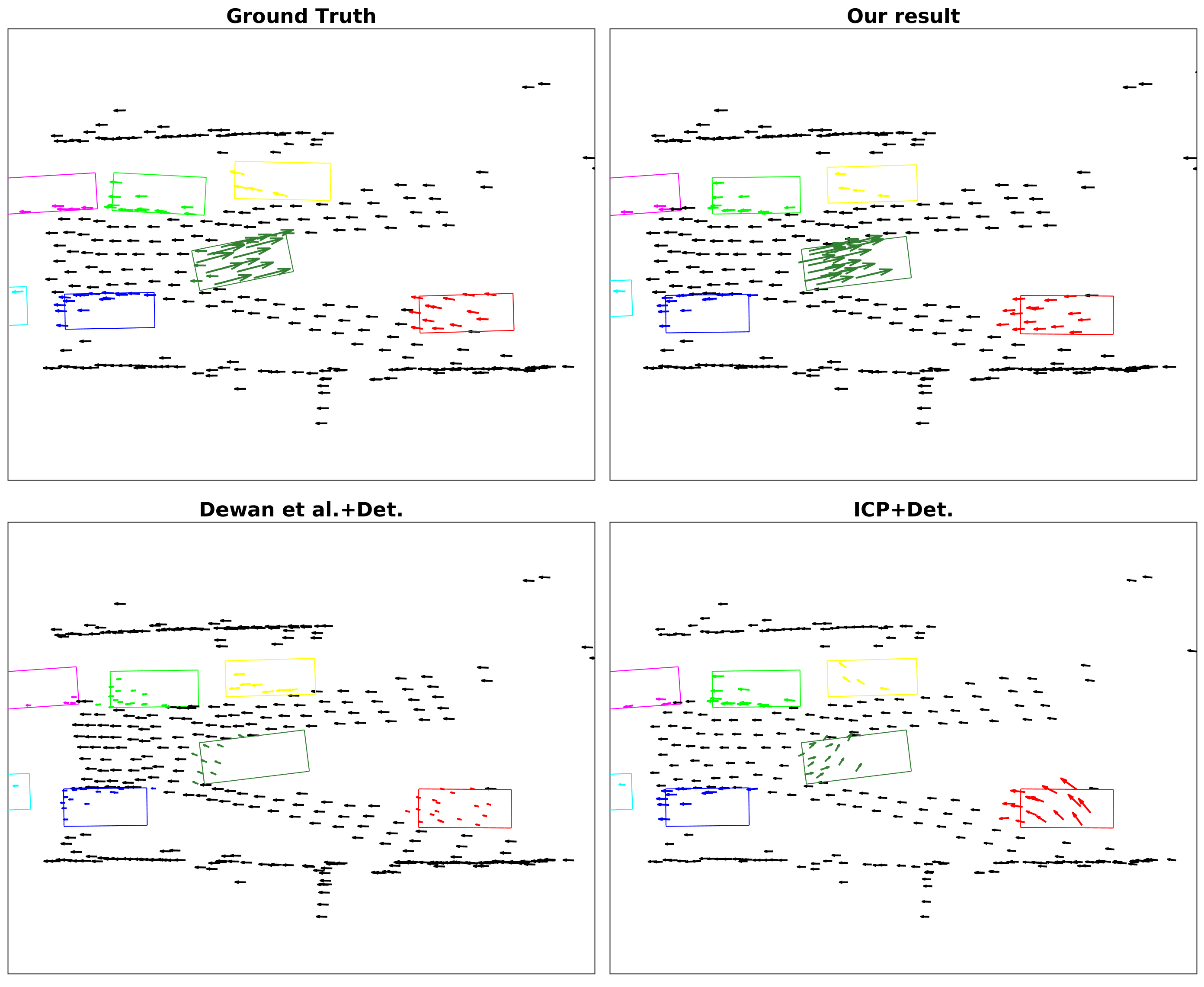}
	\end{tabular}
	\caption{
		{\capqual}}
	\label{fig:results18}
	\vspace{-0.4cm}
\end{figure*}
\begin{figure*}[ht!]
	\setlength\tabcolsep{1pt}
	\def\arraystretch{1}
	\def\imgw{\textwidth}
	\begin{tabular}{c}
		\includegraphics[width=\imgw]{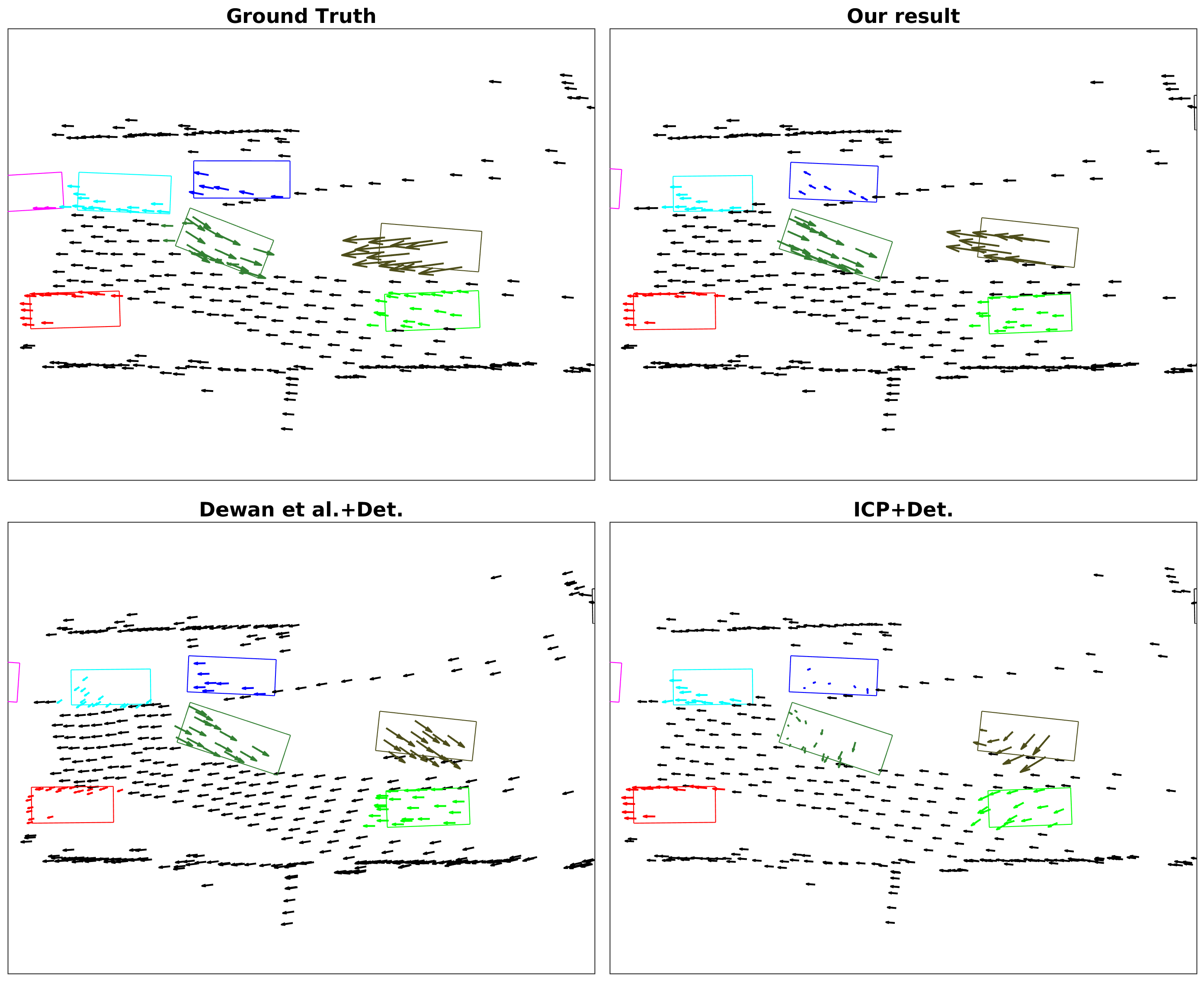}
	\end{tabular}
	\caption{
		{\capqual}}
	\label{fig:results19}
	\vspace{-0.4cm}
\end{figure*}


\end{appendices}

\end{document}